\documentclass[10pt,conference]{IEEEtran}
\IEEEoverridecommandlockouts
\usepackage{cite}
\usepackage{hyperref}
\usepackage{amsmath,amssymb,amsfonts}
\usepackage{graphicx}
\usepackage{textcomp}
\usepackage{xcolor}
\usepackage{wrapfig,lipsum,booktabs}

\def\BibTeX{{\rm B\kern-.05em{\sc i\kern-.025em b}\kern-.08em
    T\kern-.1667em\lower.7ex\hbox{E}\kern-.125emX}}

\usepackage[ruled, vlined, linesnumbered]{algorithm2e}
\usepackage{caption}
\usepackage{subcaption}
\usepackage{float}
\usepackage{multirow}
\usepackage{wrapfig}
\usepackage{comment}
\AtBeginDocument{%
  \providecommand\BibTeX{{%
    \normalfont B\kern-0.5em{\scshape i\kern-0.25em b}\kern-0.8em\TeX}}}

\begin{document}
\title{Prediction Surface Uncertainty Quantification in Object Detection Models for Autonomous Driving}

\author{\IEEEauthorblockN{Ferhat Ozgur Catak}
\IEEEauthorblockA{\textit{Simula Research Laboratory}\\
Fornebu, Norway \\
ozgur@simula.no}
\and
\IEEEauthorblockN{Tao Yue}
\IEEEauthorblockA{\textit{Simula Research Laboratory} \\
\textit{Nanjing University of Aeronautics and Astronautics}\\
taoyue@ieee.org}
\and
\IEEEauthorblockN{Shaukat Ali}
\IEEEauthorblockA{\textit{Simula Research Laboratory}\\
Fornebu, Norway \\
shaukat@simula.no}
}

\maketitle

\begin{abstract}
  Object detection in autonomous cars is commonly based on camera images and Lidar inputs, which are often used to train prediction models such as deep artificial neural networks for decision making for object recognition, adjusting speed, etc. A mistake in such decision making can be damaging; thus, it is vital to measure the reliability of decisions made by such prediction models via uncertainty measurement. Uncertainty, in deep learning models, is often measured for classification problems. However, deep learning models in autonomous driving are often multi-output regression models. Hence, we propose a novel method called PURE (Prediction sURface uncErtainty) for measuring prediction uncertainty of such regression models. We formulate the object recognition problem as a regression model with more than one outputs for finding object locations in a 2-dimensional camera view. For evaluation, we modified three widely-applied object recognition models (i.e., YoLo, SSD300 and SSD512) and used the KITTI, Stanford Cars, Berkeley DeepDrive, and NEXET datasets. Results showed the statistically significant negative correlation between prediction surface uncertainty and prediction accuracy suggesting that uncertainty significantly impacts the decisions made by autonomous driving.

\end{abstract}


\section{Introduction} \label{sec:introduction}

Deep Learning (DL) is increasingly used in autonomous driving. Existing research focuses more on increasing the accuracy of a DL model but less on studying its uncertainty \cite{9046805}. However, its quantification is essential to improve the prediction quality of the DL model.

DL is affected by two types of uncertainties. Epistemic or model uncertainty \cite{6298890} can be reduced or even avoided with sufficient and high-quality training data and optimal model configurations. The epistemic uncertainty is due to the shortage of input data; thus, the model is uncertain because it does not understand the relation between the input data and the output labels. As a result, the parameters of a neural network model are poorly optimized for unseen inputs in the prediction time. Aleatoric (also called data or irreducible) uncertainty \cite{8569814} cannot be reduced even with more data provided, e.g., sensor measurement errors. 



We propose a new method called PURE (Prediction sURface uncErtainty (PURE) to quantify prediction uncertainty of an DL-based object detection model using multiple predictions for a given image. The DL model makes multiple different bounding box predictions for a given object. PURE applies a clustering method for linking these bounding box predictions. We modified three object detection DL models (YoLo, SSD300 and SSD512) and conducted experiments with four real-world autonomous driving datasets: KITTI, Stanford Cars, Berkeley DeepDrive, and NEXET. The results of the statistical analysis showed a significant negative correlation between uncertainty and prediction accuracy. These results suggest that autonomous driving's decisions accuracy is negatively impacted by uncertainty, thus requiring new methods to deal with uncertainty.     

%



The paper is organized as follows: Section \ref{sec:related} introduces the related work. Section \ref{sec:preliminaries} describes preliminary information. Section \ref{sec:system_overview} presents our method, followed by its evaluation in Section \ref{sec:experiments}. We conclude the paper in Section \ref{sec:conclusion}.

\section{Related Work}\label{sec:related}


Object detection localizes all known object classes in an image with fixed bounding boxes and assigns a correct class label. The Monte-Carlo (MC) dropout method is used in many machine learning problems such as regression, classification, and in different domains such as medical applications \cite{leibig2017leveraging}, autonomous driving \cite{DBLP:journals/corr/abs-1811-06817}, and cybersecurity\cite{tuna2020closeness}. The MC dropout method makes predictions for an input to measure uncertainty. 



Bayesian Neural Network based MC dropout is the primary path to uncertainty quantification in DL \cite{10.5555/3104482.3104568,DBLP:journals/corr/abs-1902-02476}. An alternative is \textit{Deep Ensembles}, a non-Bayesian method for uncertainty quantification \cite{10.5555/3295222.3295387}. In \textit{Deep Ensemble}, several models are trained in parallel using random noise (with adversarial instances) in its subset part of the training dataset. The final training output is a set of independent classifiers having unique weights. The ensemble can predict the same input instance with different predictions using its individual models. Considering that Deep Ensemble requires several models, it is not suitable for single model uncertainty quantification. Therefore, we opted for the Bayesian neural networks approach.

Uncertainty quantification in deep neural networks (DNNs) is gaining attention. Softmax variance and expected entropy over multiple models have been employed for it. Gal et al. \cite{gal2016uncertainty} showed that a DNN model with prediction time dropout is equivalent to a specific variational inference on a Bayesian neural network model. The prediction model uncertainty is approximated by averaging probabilistic feed-forward MC dropout sampling during the prediction time. The approach is highly effective for practical implementation of large models and has proved useful in understanding dynamics of the networks under different testing conditions \cite{verdoja2020notes}. 
Ghoshal et al. \cite{ghoshal2020estimating} studied how Bayesian Convolutional Neural Networks can quantify uncertainty in DNN models to enhance the diagnostic performance of the human-machine alliance using a COVID-19 chest X-ray dataset. Another study \cite{schubert2020metadetect} uses uncertainty values and quality estimates to model predictive uncertainty estimation, which relies on the difference between the detection box and ground truth in images and uses standard deviation for the bounding box variables.

Michelmore et al. \cite{DBLP:journals/corr/abs-1811-06817} employed variation ratios, entropy and mutual information based uncertainty quantification methods to discover the relation between uncertainty and wrong prediction of steering angles of autonomous cars. Their results show that wrong angle decisions have significantly higher uncertainty values than correct ones. Based on this observation, they defined a threshold to determine crash risk estimation. Their DL model's inputs are camera views and the output is the steering angle, which is a single output regression model. Their method's inputs are camera views, same as for PURE. But, PURE employs existing multi-output DL models with cars' camera views to detect objects and quantify their uncertainty.

Millet et al. \cite{8793821} evaluated a set of grouping strategies for sampling based uncertainty quantification metrics. They employed density-based clustering algorithms such as HDBSCAN, Basic Sequential Algorithmic Scheme, modified the SSD model with new dropout layers, and measured object detection performance using different clustering techniques. PURE, instead, aims to build an uncertainty quantification method using density-based clustering methods.

There exist works \cite{25456,27668,23476} in modeling/measuring uncertainty for supporting the testing of cyber-physical system (CPSs). These works however do not focus on DL models. However, they can be integrated with PURE for supporting uncertainty quantification of CPSs with DL models applied. 


\section{Preliminaries}\label{sec:preliminaries}

\subsection{Object Detection}
Object detection is an algorithmic approach in deep learning related to computer vision and image processing that deals with detecting a specific label's objects in digital images and videos. It is widely used in autonomous driving, face detection, plate recognition, and medical imaging. 
Classical object detection techniques use handcrafted features and lightweight machine learning models \cite{8627998}. Thus, it is challenging for them to predict whether an object is likely to be the one of interest, specially for image recognition. Powerful neural network architectures such as SSD, YoLo are now available to address such challenges \cite{mattson2020mlperf}. 

\subsection{Uncertainty in Deep Learning}

Gal et al. \cite{gal2016uncertainty} (also see Section~\ref{sec:related}) proposed an MC Dropout approach adding intermediate dropout layers to a model without changing the numbers of neurons and layers of existing object recognition models. As an ensemble approach, for each of its single ensemble model, the system drop outs different neurons in the network's each layer according to the dropout ratio in the prediction time. The predictive mean is the average of the predictions over dropout iterations, $T$, and is used as the final inference, $\hat{y}$, for the input instance $\mathbf{x}$ in the dataset. The overall prediction uncertainty is approximated by finding the entropy and the probabilistic feed-forward MC dropout sampling variance during prediction time. The prediction is defined as follows \cite{gal2016uncertainty}:

\begin{equation}
 p(\hat{y}=c|\mathbf{x},\mathcal{D}) \approx \hat{\mu}_{pred}  = \frac{1}{T} \sum_{y \in T} p(\hat{y}|\mathbf{\theta},\mathcal{D})
\end{equation}
where $\theta$ is the model weights, $\mathcal{D}$ is the input dataset, $T$ is the number of predictions of the MC dropouts, and $\mathbf{x}$ is the input sample. The label of input sample $\mathbf{x}$ can be estimated with the mean value of MC dropouts predictions $p(\hat{y}|\mathbf{\theta},\mathcal{D})$, which will be done $T$ times. In the prediction time, random neurons in each layer are dropped out, based on $p$, from the base DNN model to create another model. As a result, $T$ different DNN models can be used to predict the input instance's class label and uncertainty quantification of the overall prediction.

\subsection{Uncertainty Sources}
\begin{figure}[!htbp]
    \centering
    \includegraphics[width=0.9\linewidth]{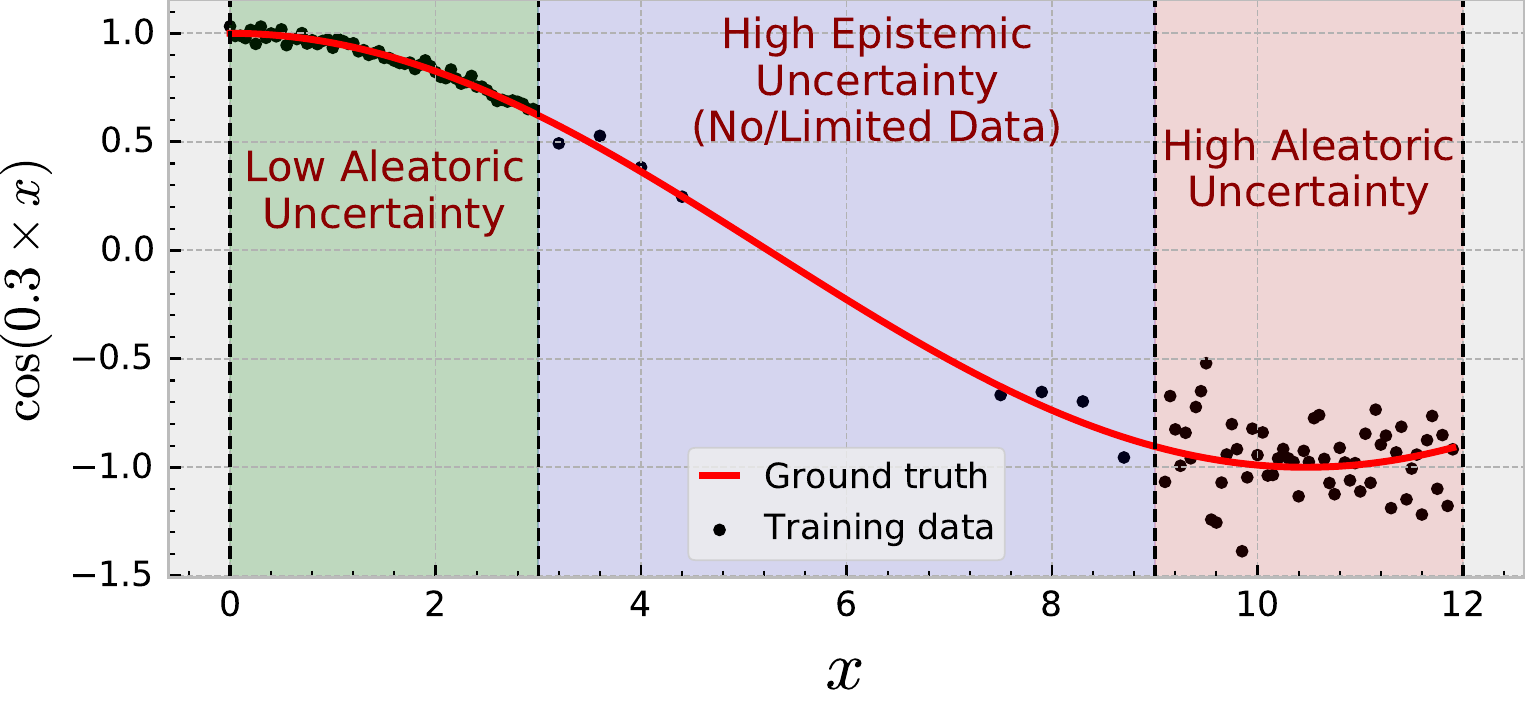}
    \caption{Illustrating uncertainty types with a regression model.}
    \label{fig:regression_uncertainty_regions}
\end{figure}

DNN models are affected by epistemic \cite{6298890} and aleatoric \cite{8569814} uncertainties, as briefly explained in Section \ref{sec:introduction}. Epistemic uncertainty should be eliminated by taking additional training data and making optimal model configurations (i.e. number of layers, activation functions, optimization) suitable for the problem domain. Aleatoric uncertainty cannot be reduced even with more data. Measurement errors (e.g., in sensor inputs) are a common source of confusion in neural networks because there is no guarantee that the resulting prediction will be correct. Figure \ref{fig:regression_uncertainty_regions} explains the differences between the two uncertainty types for a simple regression model.

\subsection{Uncertainty Quantification Methods}

A typical DNN model predicts a single output for a regression problem or a softmax probability of each label for a classification problem. To quantify a single prediction's uncertainty, the model should produce multiple outputs for the single input instance. An uncertainty quantification metric is a scalar value, measuring the reliability of the prediction for an input instance. For regression models, the only uncertainty quantification metric is the variance of the predictions:

%
%
%
%

\begin{equation}
    S^2 = \frac{\sum_{i=1}^T(y_i - \overline{y})}{T-1}
\end{equation}
where $y_i$ is the $i$th prediction of the input instance $\mathbf{x}$, and $\overline{y}$ is the average of all the predictions, $y = [y_1, \cdots, t_T]$. Such prediction variance based uncertainty quantification is only applicable for single output regression models. For multi-output regression problems, we need another approach. Typically, an object detection model (e.g., Yolo and SSD) outputs a rectangle for each predicted object from an two-dimensional input image. The rectangle is defined in 4 corners in the image. We propose \texttt{prediction surface}, a novel uncertainty quantification method for multi-output regression models. In this approach, the object detection model makes $T$ different predictions with the MC dropout method for the input image $X$. For each predicted object, $\mathcal{O}_i$, in the image $X$, the predicted corners locations ($x, y$) are calculated using the convex hull method.

\section{The PURE Method}\label{sec:system_overview} 
\begin{figure}[!htbp]
\centering
	\includegraphics[width=1.0\linewidth]{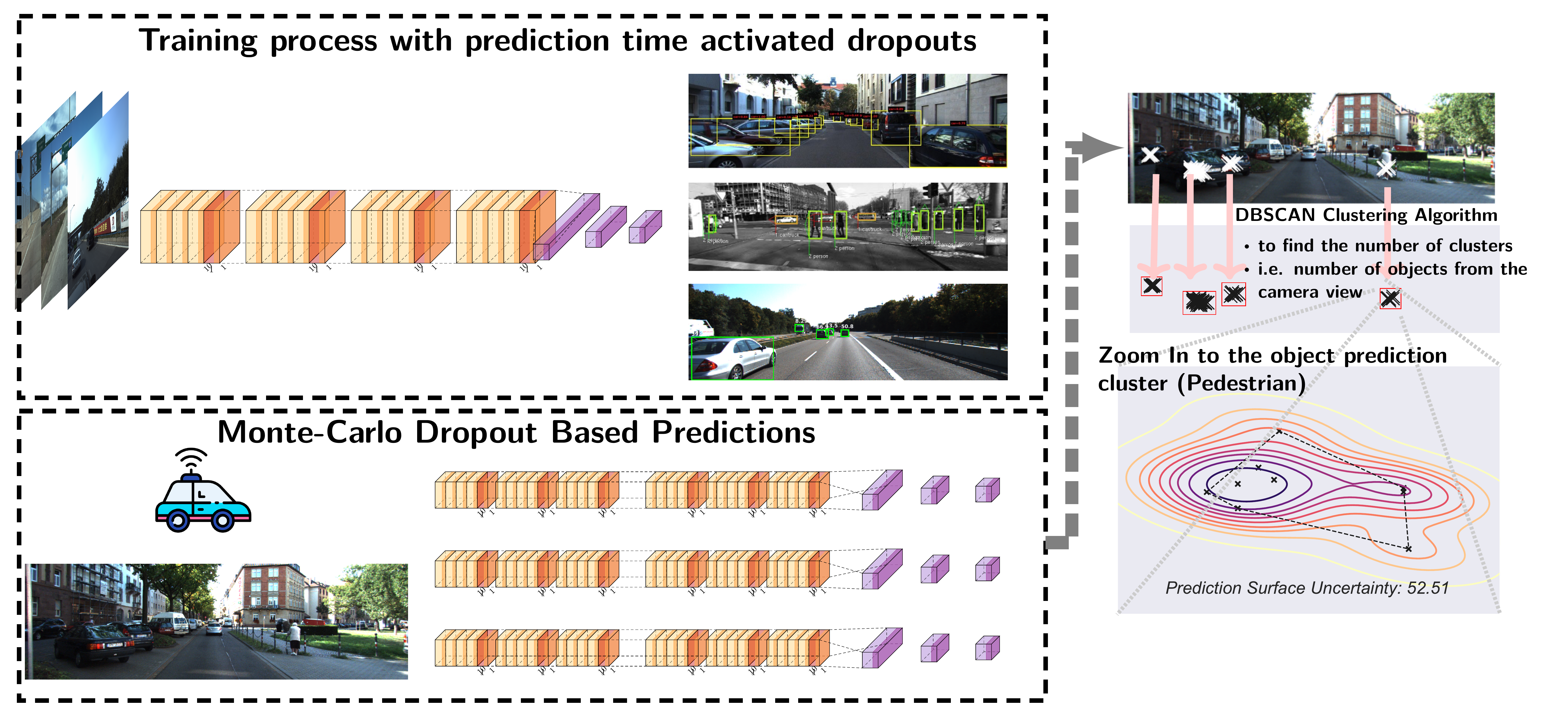}
	\caption{Overview of PURE.}
	\label{fig:system-overview}
\end{figure}

The overview of PURE is given in Figure \ref{fig:system-overview}. First, an object detection model is trained with an annotated (i.e. bounding boxes) image dataset. We then modify such pre-trained object detection models (e.g., YoLo or SSD) to be compatible with the MC dropout method. To do so, we inject the prediction-time activated dropout layers to the existing models to create small variations of the current predictive models to find the overall predictions' uncertainty. The modified models discover underlying objects within the testing images and predict corresponding object locations in two dimensions.


\subsection{Problem Formulation}

Machine learning models extract patterns from their training data. A typical object detection task, thus, detects all available objects from the camera scene. However, the problem is that the number of objects in a camera image can be quite large, and their sizes are quite variable. Also, datasets used in the training of object detection models are usually labeled manually and non-systematically. Thus, errors occur in predicting objects in a system, such as in the autonomous car's DNN model. Thus, our motivation is to assist the object detection predictions by quantifying each object's uncertainty in a camera scene.

For an object detection problem, the object location can be defined as the start-point ($x_1, y_1$) and the endpoint ($x_2, y_2$) of the rectangle. In MC dropout-based object detection, it is challenging to quantify the uncertainty of a model's multiple predictions. The prediction has four different continuous values ($x_1, x_2, y_1, y_2$) to define detected object boundaries. The standard deviation is the only quantification method for single-output regression models to the best of our knowledge \cite{weiss2021uncertaintywizard}. In object detection with autonomous cars' camera views, there can be an unknown number of objects (e.g., pedestrians, cars, traffic signs), and each has 4 location information defining its boundary. Moreover, the number of objects in a scene cannot be quantified easily, as each object cannot be detected in each Bayesian inference.

The MC dropout-based uncertainty quantification method makes multiple predictions with randomly dropped neurons from the same model. In a highly certain model of having low epistemic uncertainty, the distances between multiple predictions should be very close. Ideally, all the predictions should be exactly at the same points. Moreover, there can be a different number of object predictions (e.g. several cars, pedestrians) in each MC prediction for a single camera view. Therefore, the MC dropout based model can produce predictions with different locations for each object in a single image.

Thus, the problems in object detection from car camera views are: detecting the number of objects within all prediction results in Bayesian Neural Networks, and quantifying prediction uncertainty for detected object boundaries.

\subsection{Algorithms}
Algorithms \ref{alg:overall_1}-\ref{alg:overall_2} outline the top-level steps implemented in PURE: 1) the pre-trained object detection model modified with prediction time activated dropout layers, and 2) density-based object recognition grouping from the modified object detection model's predictions with MC dropout. 

Algorithm \ref{alg:overall_1} takes the object detection model $h$ as the input, lets it make $T$ times MC dropout predictions for a given image, and records all predictions and their centers. 


Since in an prediction of the model, there are several predicted rounding-boxes for a single ground-truth object. To assign each prediction to an object, clustering is needed \cite{s20164424}, which is implemented in Algorithm~\ref{alg:overall_2}. 
\begin{algorithm}[htbp!] \footnotesize
\DontPrintSemicolon 
\KwIn{$X \in \mathbb{R}^{m \times n}$: camera view \\ \hspace{13pt} $h$: Object detection model \\ \hspace{13pt} $T$: MC dropout size}
\KwOut{Predictions $center\_lists$}
$center\_lists \gets \emptyset $ \;
\tcc{$T$ predictions using the object detection model $h$}
\ForEach {$t \in T $} 
{
     \tcc{Predict objects with surrounding boxes from image $X$}
     $VBoxes \gets h(X)$ \; 
     \ForEach{$box \in VBoxes$} 
     {
        \tcc{Get the corner locations of each surrounding box}
        $x_1, y_1 = box.xmin, box.ymin, $ \; %
        $x_2, y_2 = box.xmax, box.ymax$ \; %
        \tcc{Find the width and height of the predicted object}
        $width, height = x_2 - x_1, y_2 - y_1$ \;
        $x_{center}, y_{center} \gets x_1 \frac{width}{2}, y_1 + \frac{height}{2}$ \;
        $center\_lists \gets UpdateArchive(x_{center}, y_{center}) $ \;
     }
}
\Return{$center\_lists$}
\caption{MC dropout based object predictions}\label{alg:overall_1}
\end{algorithm}
Algorithm \ref{alg:overall_2} clusters the object predictions' centers to label objects (e.g. cars, pedestrians, signs) (line 2). We use the DBSCAN clustering algorithm to find objects \cite{10.1145/3068335}. DBSCAN is a parameter-free clustering algorithm, which tries to find densities based on radius, $\epsilon$, and minimum samples in the radius. We set $\epsilon$ as 100 and the minimum samples as 3. 
Then, we calculate the area of the predicted corners with convex hull (smallest convex shape enclosing all points) (lines 3-8). Four different locations (i.e. corners of the box) define an object from the predictions: $(x_1,y_1)$, $(x_1,y_2)$, $(x_2,y_1)$, $(x_2,y_2)$. We average all the points' area to get the uncertainty of the input image $X$.

\begin{algorithm}[htbp!] \footnotesize
\DontPrintSemicolon 
\KwIn{Predictions $center\_lists$, \\ \hspace{13pt} $\epsilon$: radius of DBSCAN}
\KwOut{Uncertainty of predictions, $\mathcal{U}$}
$area \gets \emptyset$ \;
\tcc{Predictions are completed. Find object clusters from the $center_list$}
$Clusters \gets DBSCAN(center\_list, \epsilon)$ \;
\ForEach{cluster $c \in Clusters$}
{
    \tcc{Calculate the area of each corner of the predictions in cluster $c$}
    area $x_1 \gets ConvexHull(center\_lists[c].x_1)$ \;
    area $y_1 \gets ConvexHull(center\_lists[c].y_1)$ \;
    area $x_2 \gets ConvexHull(center\_lists[c].x_2)$ \;
    area $y_2 \gets ConvexHull(center\_lists[c].y_2)$ \;
    \tcc{Calculate the cluster's (i.e. the single object's) uncertainty}
    $area \gets$ UpdateArchive(Mean(area $x_1$, area $y_1$, area $x_2$, area $y_2$))\;
}
\tcc{Average uncertainty of all detected objects from all $T$ predictions}
$\mathcal{U} \gets Mean(area)$ \;
\tcc{Return calculated uncertainty $\mathcal{U}$ for the predictions from the image $X$}
\Return{$\mathcal{U}$}
\caption{DBSCAN clustering algorithm based clustering of predictions and uncertainty quantification}\label{alg:overall_2}
\end{algorithm}

\begin{figure}[!htbp]
    \centering
    \begin{subfigure}[b]{1.0\linewidth}
         \centering
         \captionsetup{justification=centering}
         \includegraphics[width=1\linewidth]{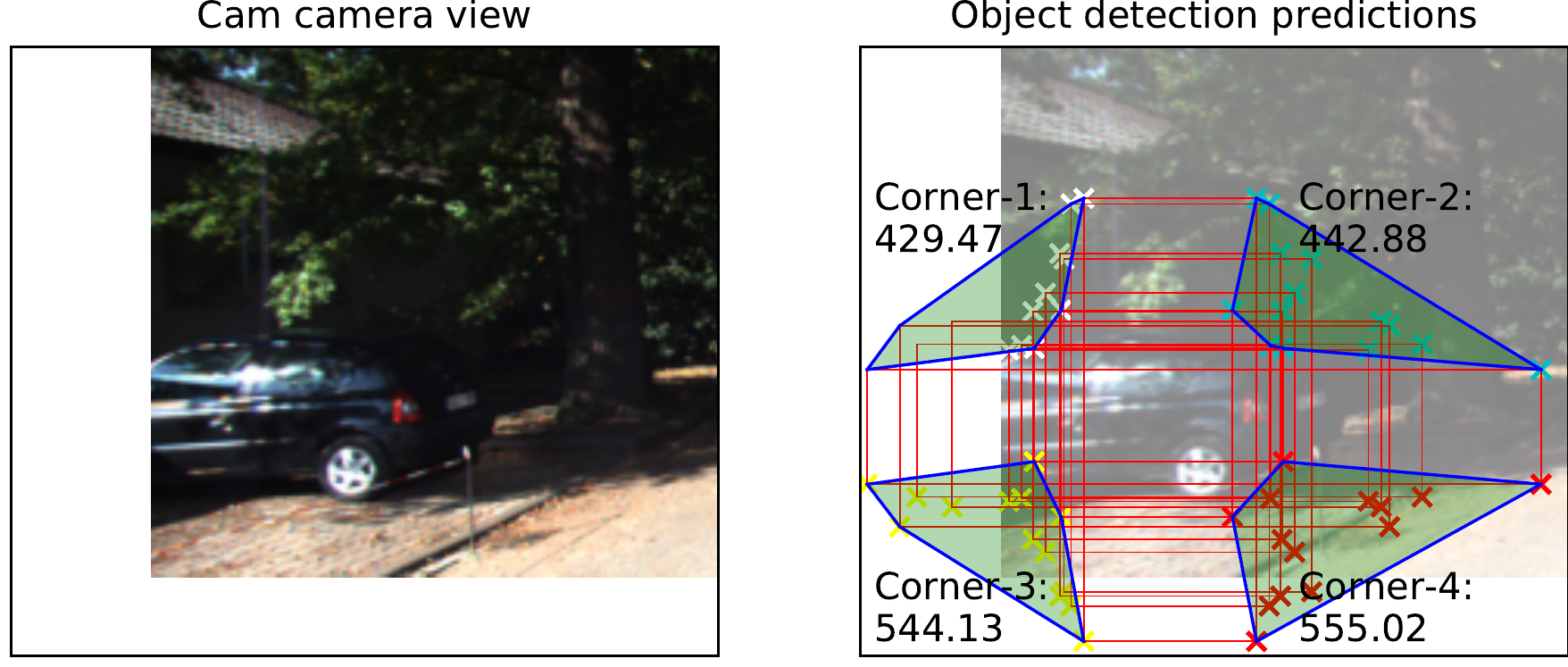}
         \caption{High uncertainty}
	    \label{fig:high_uncertainty}
     \end{subfigure}\hfill
     \begin{subfigure}[b]{1.0\linewidth}
         \centering
         \captionsetup{justification=centering}
         \includegraphics[width=1\linewidth]{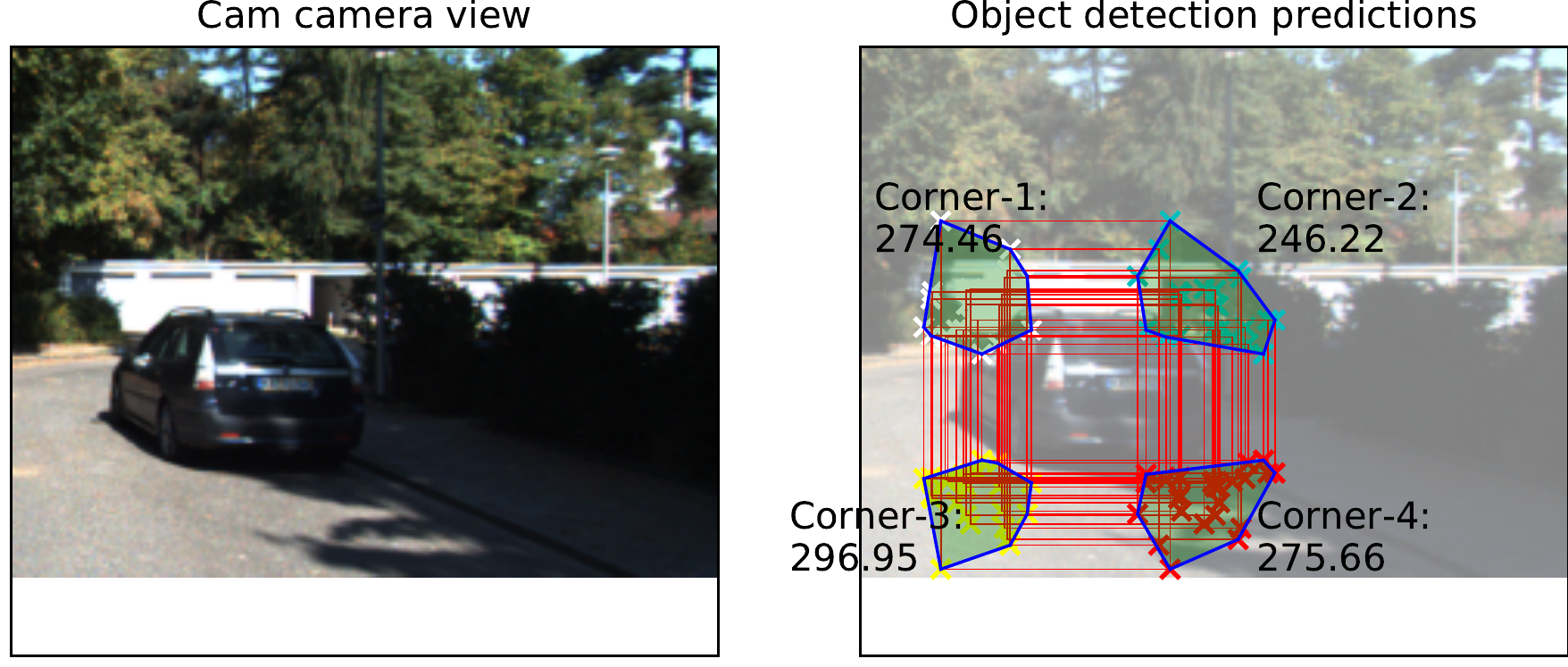}
         \caption{Low uncertainty}
	    \label{fig:low_uncertainty}
     \end{subfigure}\hfill
	\caption{Illustrating high and low uncertainty of object detection predictions. }
	\label{fig:low_high_uncertainty}
\end{figure}

Figure \ref{fig:low_high_uncertainty} shows the low and high uncertain predictions from the modified YoLo object detection model. Figure \ref{fig:high_uncertainty} shows the highly uncertain predictions for detecting a car in the camera view. In this image, the variability of the prediction boxes' corners is quite high; the perception of the object detection model has high uncertainty. Figure \ref{fig:low_uncertainty} shows the low uncertainty predictions for another image. The predicted corners are closer to each other. 

We implemented PURE in Python with the Tensorflow deep learning library. The code is available from GitHub \footnote{\url{https://github.com/Simula-COMPLEX/pure}}.

\section{Experiments}\label{sec:experiments}
In this section, we first present the research questions (RQs), followed by the object detection models and datasets employed in the experiment. We also present experiment metrics and procedure, followed by results and threats to validity. 
\subsection{Research Questions}\label{RQ}
[\textbf{RQ1}] Is the prediction surface an effective uncertainty quantification method for object detection predictions? This RQ assesses the effectiveness of PURE. 
[\textbf{RQ2}] Is there any correlation between prediction surface uncertainty and the object detection performance of DNN models? This RQ tests the hypothesis that the prediction performance of a DNN model decreases with the increase in the uncertainty in images used for object detection.

\subsection{Object Detection Models}\label{object detection models}
We chose three DNN-based object detection models: YoLo \cite{redmon2018yolov3}, SSD300 and SSD512 \cite{10.1007/978-3-319-46448-0_2} since they are publicly available. We modified them to ensure their compatibility with the MC dropout approach. More specifically, for SSD300 and SSD512, we added dropout layers on the convolutional layers of the VGG16 component with varying dropout probabilities, $p\in[0.1, \cdots , 0.5]$. For YoLo, we added dropout layers to the last three convolutional layers (i.e. between 80-82, 92-94, and 99-106). We used the open-source code of Tensorflow versions of Python code of all the model and added the dropout layers using Tensorflow.

\subsection{Datasets}\label{datasets}
We employed the KITTI \cite{Menze2015CVPR}, Stanford Cars \cite{KrauseStarkDengFei-Fei_3DRR2013}, Berkeley DeepDrive \cite{yu2020bdd100k}, and NEXET datasets (Table \ref{tab:ds_detailt}).

\begin{table}[htbp!]
    \centering
    \caption{Characteristics of the employed datasets}
    \label{tab:ds_detailt} 
    \begin{tabular}{c|c|c}
    \hline
        \textbf{Dataset} & \textbf{Number of images} & \textbf{Resolution} \\ \hline \hline
        KITTI & 7,481 & $1224 \times 370$ \\
        Stanford Cars & 8,144 & $300 \times 200$ \\ 
        Berkeley DeepDrive & 50,004 & $1280 \times 720$\\
        NEXET & 44,527 & $1280 \times 720$ \\ \hline
    \end{tabular}
    
\end{table}

\subsection{Object Detection Performance Metrics}\label{metrics}
An object detection model predicts the bounding box's coordinates around an object when it is present in an input image (i.e. localization) \cite{electronics10030279}. Thus, we compare the coordinates of the ground truth and predicted bounding boxes. The central concept in object detection is Intersection over Union (IoU), also called Jaccard Index, which calculates the junction (the overlapping area) over the ground truth box $B_{gt}$ and the predicted bounding box $B_p$ over their union \cite{padillaCITE2020}, which is:


\begin{equation}
    IoU = J(B_p, B_{gt}) = \frac{area(B_p \cap B_{gt} )}{area(B_p \cup B_{gt})}
\end{equation}



For a perfect object detection model, IoU is always 1 meaning that the DNN model exactly predicts all ground-truth objects with the exact localization points. 
A threshold value, $t$, is needed to convert IoU to be compatible with Precision, Recall and $F_1$ evaluation metrics. If $IoU \ge t$ then the DNN model's object prediction is correct, If $IoU < t$ then the model's prediction is considered as wrong.



\subsection{Experimental Procedure}\label{experimental procedure}
The procedure has three phases.
\textbf{First}, we injected prediction time activated dropout layers after each convolutional layer of SSD300 and SSD512, and the last three convolutional layers of YoLo. To inject new dropout layers, we need two components: (1) the model structure in the Python language, and (2) the model weights in Hierarchical Data Format version 5. We modified the original models' architecture, i.e., their corresponding Python implementations, by adding new dropout layers. The optimization methods, activation functions, and the loss functions of the original models remain unchanged. Afterwards, we loaded the trained weights into the modified versions of the models. Files of the model weights contain only the input weights of each neuron in each model.
\textbf{Second}, we ran MC dropouts 20 times, $T=20$, on each dataset and each model to quantify the prediction surface uncertainty. We set $t=0.5$ as the threshold value. Dropout ratios used were $[0.1 \cdots, 0.5]$. At the end of the executions, we collected corner values for each object of each image as well as the clusters that belong to the objects in the images. 
\textbf{Third}, we bench-marked the MC dropout based models' detection with the original models, i.e. unmodified versions.




\subsection{Results}\label{results}
\subsubsection{RQ1}
Table \ref{tab:model_detection_perf} shows the results of each model for each dataset. We obtained the best results from the Stanford dataset, i.e., the highest average IoU, Recall and $F_1$, but not Precision. The reason is that there is only one car object in each picture of this dataset, i.e., ground-truth labels contain only one object in the images. However, there are other objects in the dataset, such as pedestrians, other cars without ground-truth labels. The models also detect other objects. This results in a decrease in precision for the Stanford cars dataset. 

\begin{figure}[htbp!]
    \centering
    \includegraphics[width=1\linewidth]{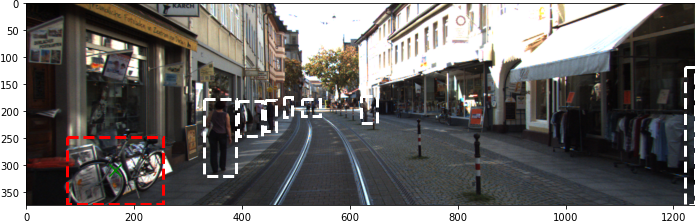}
    \caption{Illustrating YoLo's object detection predictions for KITTI. }
    \label{fig:yolo_recall_problem}
\end{figure}

The worst results were obtained on the Berkeley dataset due to the reason that each picture in this dataset has too many objects leading to the low performance of the models. According to the table, precision values are generally high, and recall values are generally low. For this dataset's models, the detected (i.e., predicted) objects (positive samples) are generally correct. Moreover, the recall values are low since the models cannot detect other objects (e.g. other cars, pedestrians) on the image. To summarize, if the model says that it has found an object, this detection is correct, but the model cannot find every object in the image. Figure \ref{fig:yolo_recall_problem} shows an example of the incompatibility of the existing object detection models with the publicly available datasets. The white boxes show the ground-truth and the red box shows the prediction. In the figure, there are 7 labelled objects and the YoLo model only detects one object in the figure. Because of this detection problem, values of the \textit{Recall} metric are very low for all the predictions. 

\begin{figure*}
    \centering
    \includegraphics[width=0.8\linewidth]{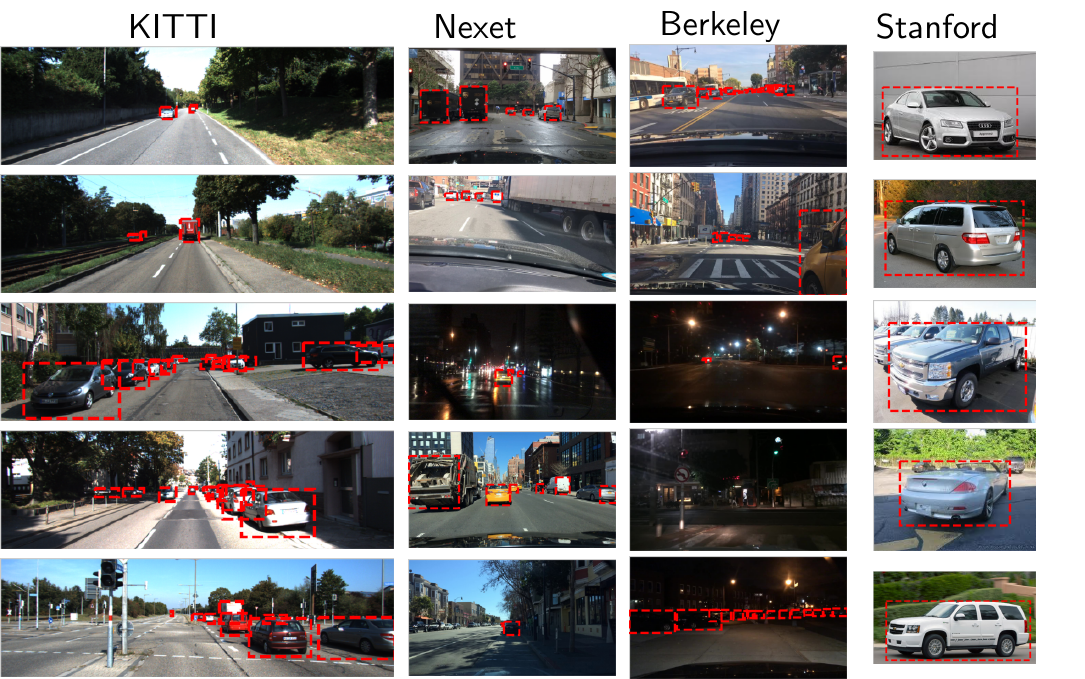}
    \caption{Examples of the datasets with ground truth (red rectangles in the images).}
    \label{fig:ds_defs}
\end{figure*}

Figure \ref{fig:ds_defs} shows five examples from each dataset. The red rectangles are the ground-truth information that the related picture has. The Stanford dataset usually contains only one object and is very easy to detect. Besides, some pictures of the Stanford dataset (e.g. pictures 3 and 4, counting from the top to down) may contain more than one objects. However, this dataset contains only one ground-truth information for each picture. In this case, even if IoU (for the detected object) is high, the precision may be low due to a large number of detected objects. KITTI is the most challenging dataset in terms of object detection. Each picture contains a large number of objects and object sizes are relatively small when compared to the Stanford dataset. Ground-truth information is quite detailed and includes small objects like cars and pedestrians in the images.



\begin{table}[htbp!]
    \centering 
    \caption{Object detection prediction results of all the models for all the datasets}
    \label{tab:model_detection_perf} \footnotesize
    \begin{tabular}{c|p{1.5cm}|c|c|c|c}  \hline
        Model & Dataset & Avg. IoU & Precision & Recall & $F_1$ \\ \hline \hline
         \multirow{4}{*}{SSD300} & KITTI & 0.5506 & 0.9065 & 0.2192 & 0.3558 \\
          & Berkeley& 0.6600 & 0.4881 & 0.5000 & 0.4940 \\
          & NEXET & 0.8483 & 0.6111 & 0.3548 & 0.4489 \\
          & Stanford & 0.8984 & 0.637 & 1.0000 & 0.7783  \\ \hline
         \multirow{4}{*}{SSD512} & KITTI & 0.7313 & 0.8437 & 0.3576 & 0.5023 \\
          & Berkeley & 0.7053 & 0.3694 & 0.5409 & 0.4390 \\
          & NEXET & 0.8341 & 0.6875 & 0.4583 & 0.5500 \\ 
          & Stanford & 0.9469 & 0.6608 & 1.000 & 0.7957 \\ \hline
         \multirow{4}{*}{YoLo} & KITTI & 0.4453 & 0.8916 & 0.1946 & 0.3195 \\
          & Berkeley & 0.5968 & 0.2115 & 0.1182 & 0.1517 \\
          & NEXET & 0.6378 & 0.7037 & 0.0902 & 0.1600 \\ 
          & Stanford & 0.8183 & 0.7931 & 0.7931 & 0.7931 \\
         \hline
    \end{tabular}
\end{table}

\begin{figure}[htbp!]
    \centering
    \begin{subfigure}[b]{0.45\linewidth} \centering
    \includegraphics[width=0.9\linewidth]{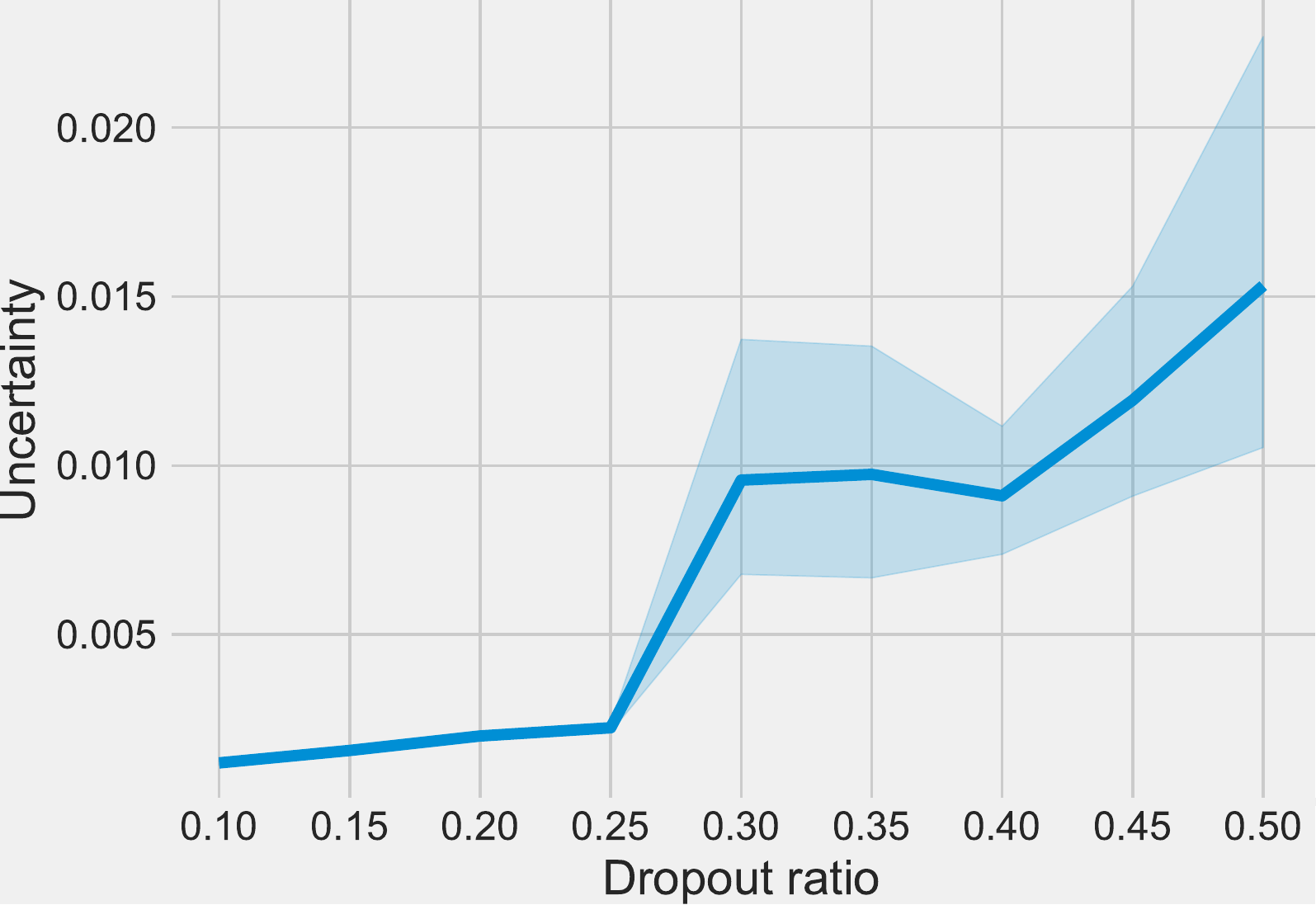}
    \caption{Stanford}
    \label{fig:ssd512-stanford_cars-unc.pdf}
    \end{subfigure}
    \begin{subfigure}[b]{0.45\linewidth} \centering
    \includegraphics[width=0.9\linewidth]{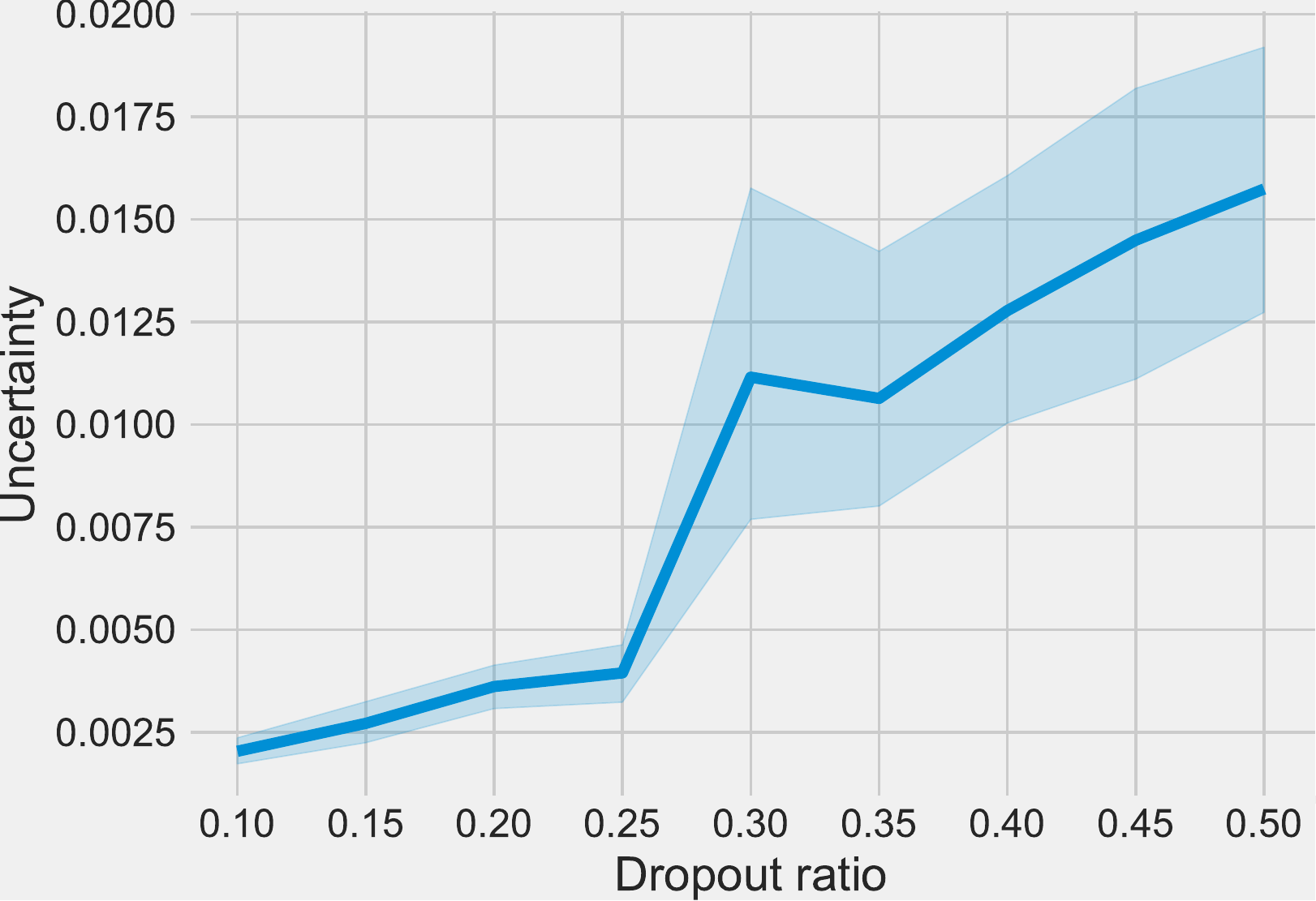}
    \caption{Berkeley}
    \label{fig:ssd512-nexet-unc.pdf}
    \end{subfigure}
    \begin{subfigure}[b]{0.45\linewidth} \centering
    \includegraphics[width=0.9\linewidth]{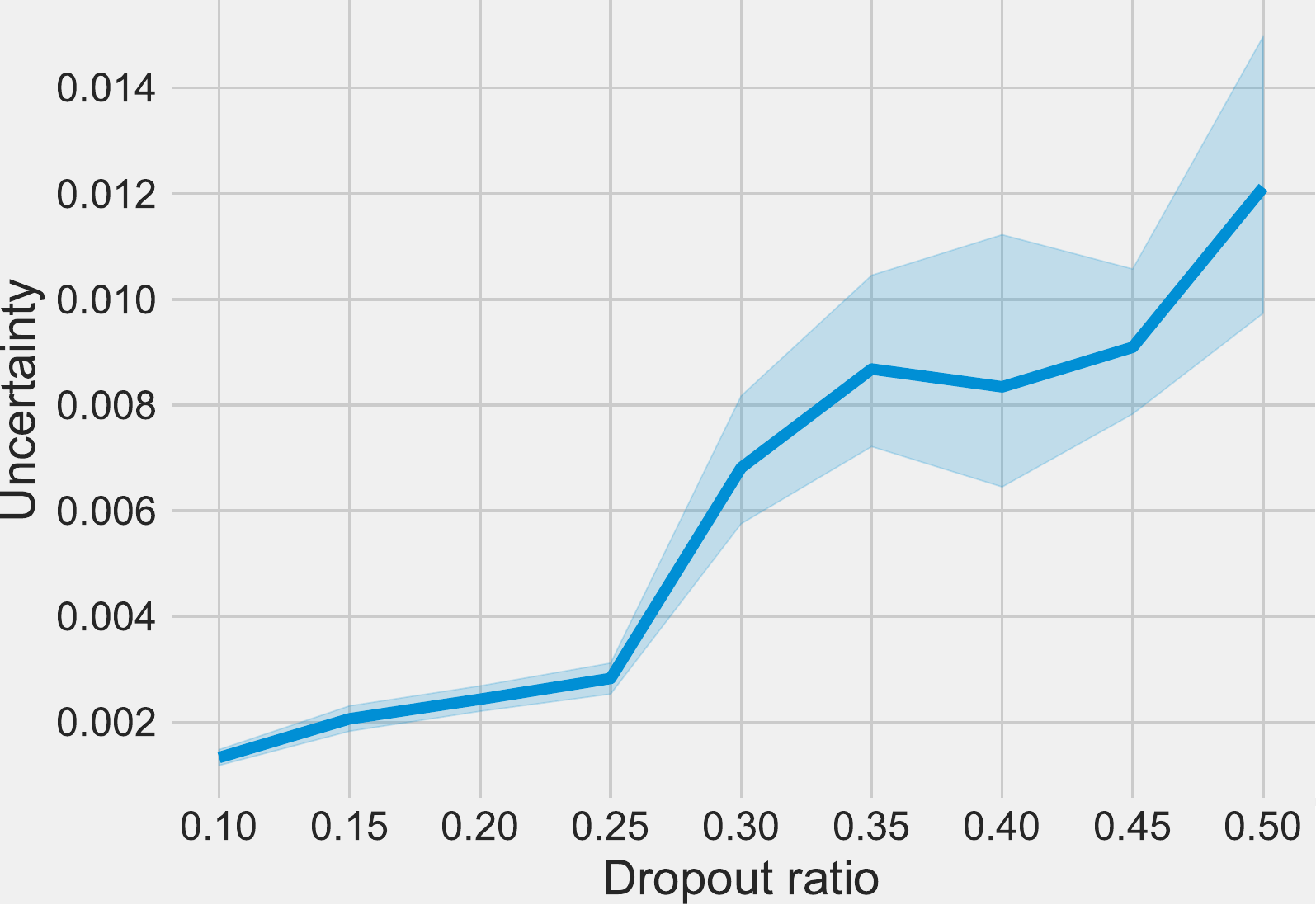}
    \caption{NEXET}
    \label{fig:ssd512-nexet-unc.pdf}
    \end{subfigure}
    \begin{subfigure}[b]{0.45\linewidth} \centering
    \includegraphics[width=0.9\linewidth]{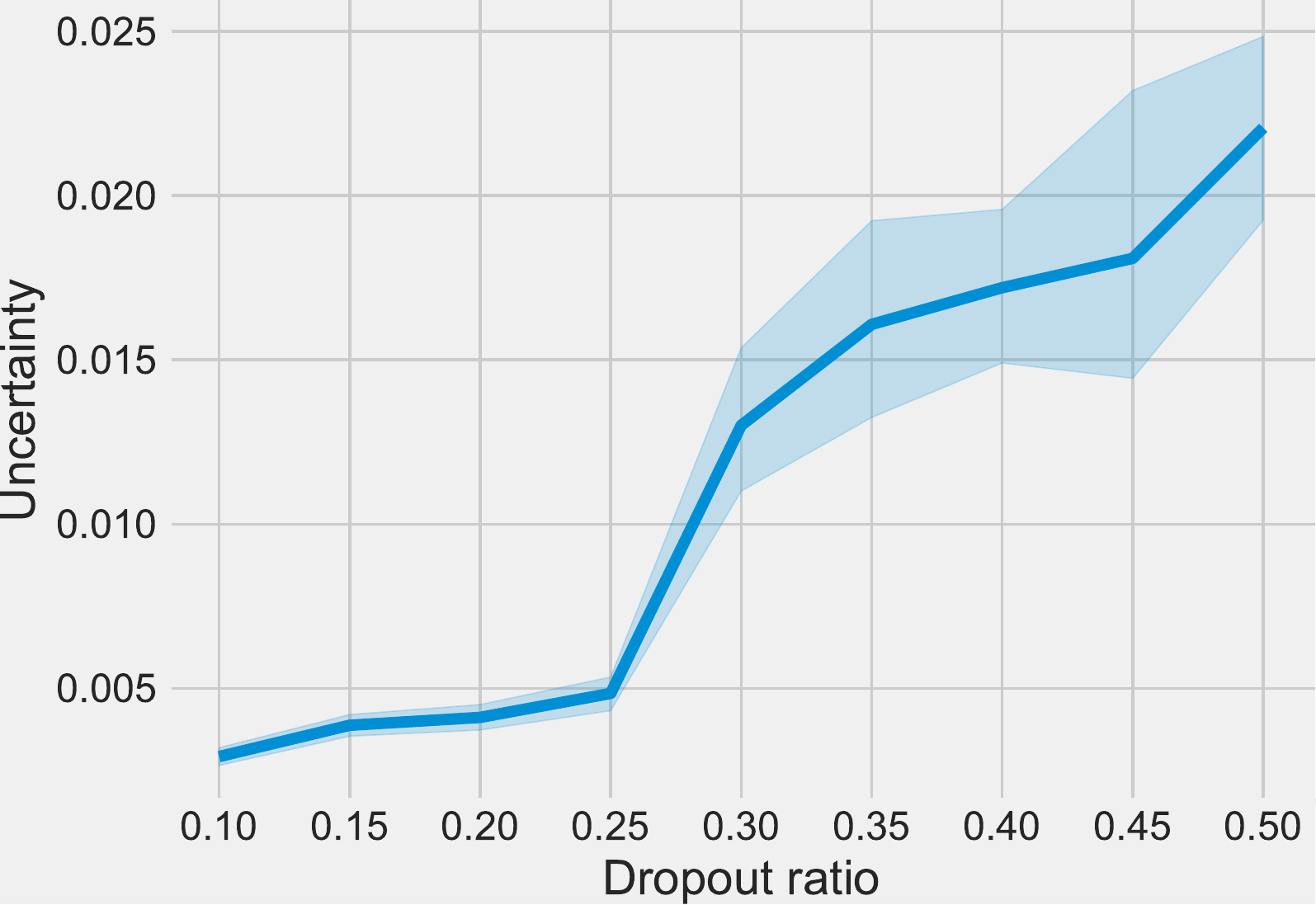}
    \caption{KITTI}
    \label{fig:ssd512-kitti-unc.pdf}
    \end{subfigure}
    \caption{SSD512 model uncertainty changes with different $p$.}
    \label{fig:ssd512-unc}
\end{figure}

\begin{figure}[h]
    \centering
    \begin{subfigure}[b]{0.48\linewidth} \centering
    \includegraphics[width=0.9\linewidth]{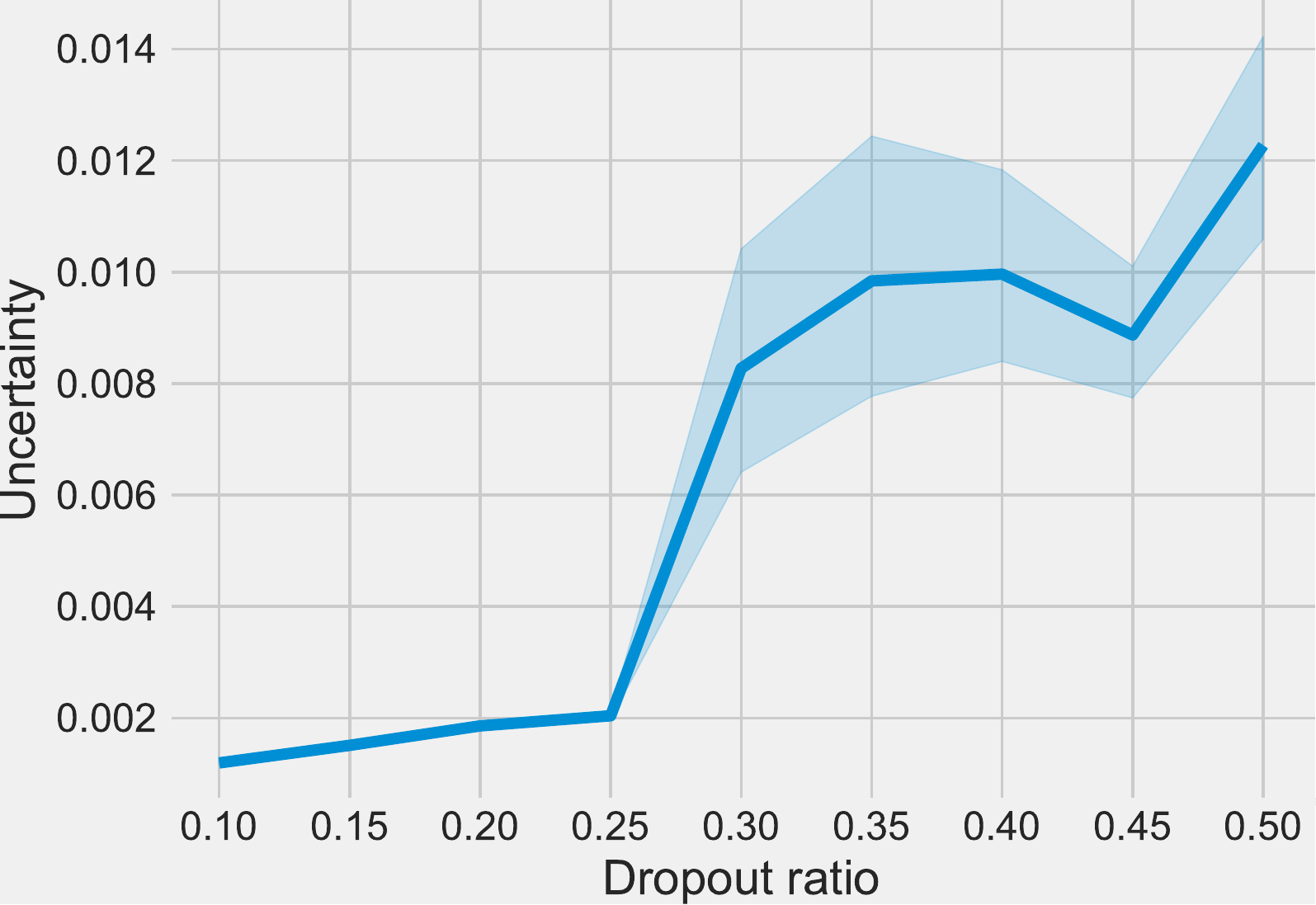}
    \caption{Stanford cars}
    \label{fig:ssd300-stanford_cars-unc.pdf}
    \end{subfigure}
    \begin{subfigure}[b]{0.48\linewidth} \centering
    \includegraphics[width=0.9\linewidth]{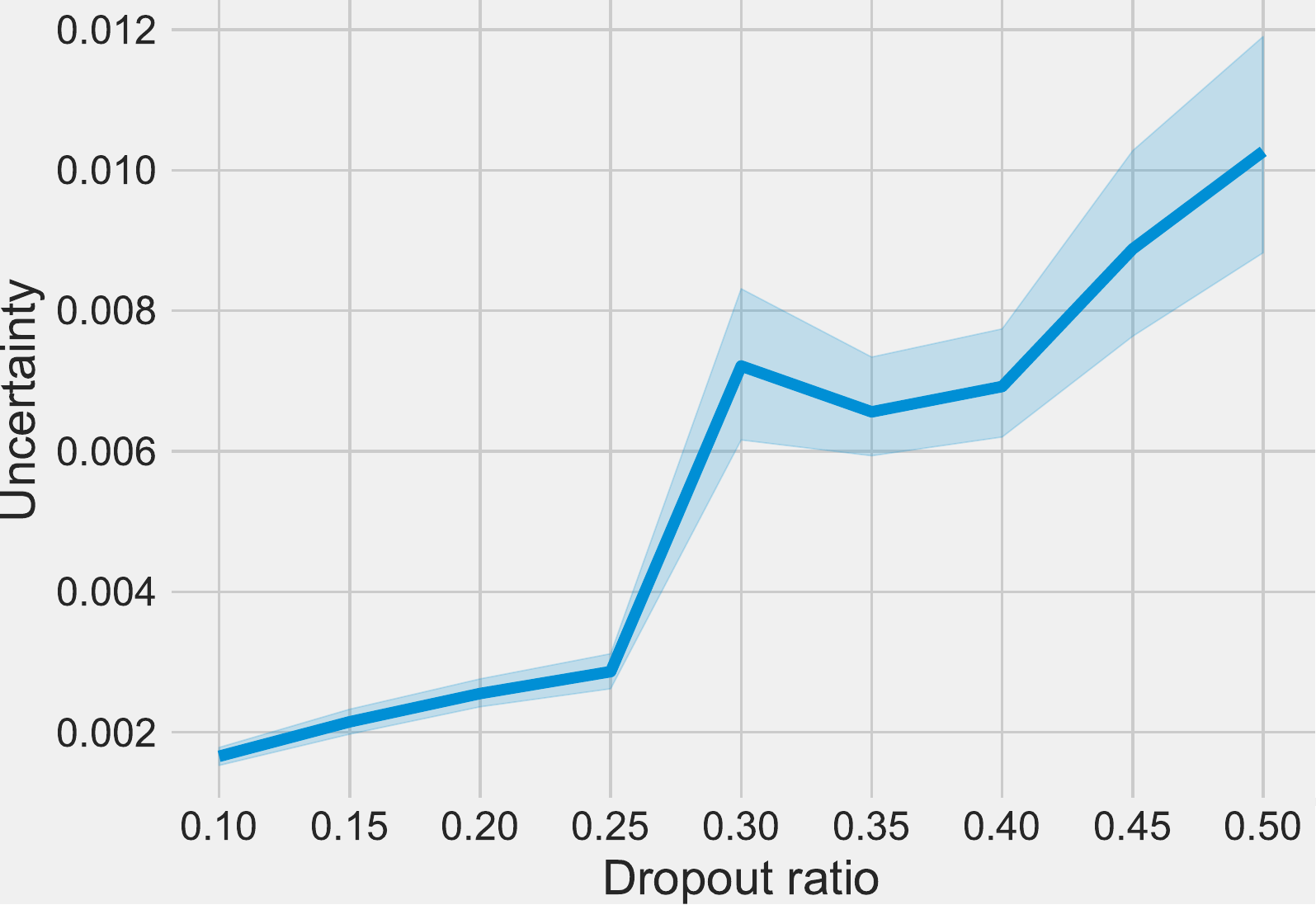}
    \caption{Berkeley DeepDrive}
    \label{fig:ssd300-nexet-unc.pdf}
    \end{subfigure}
    \begin{subfigure}[b]{0.48\linewidth} \centering
    \includegraphics[width=0.9\linewidth]{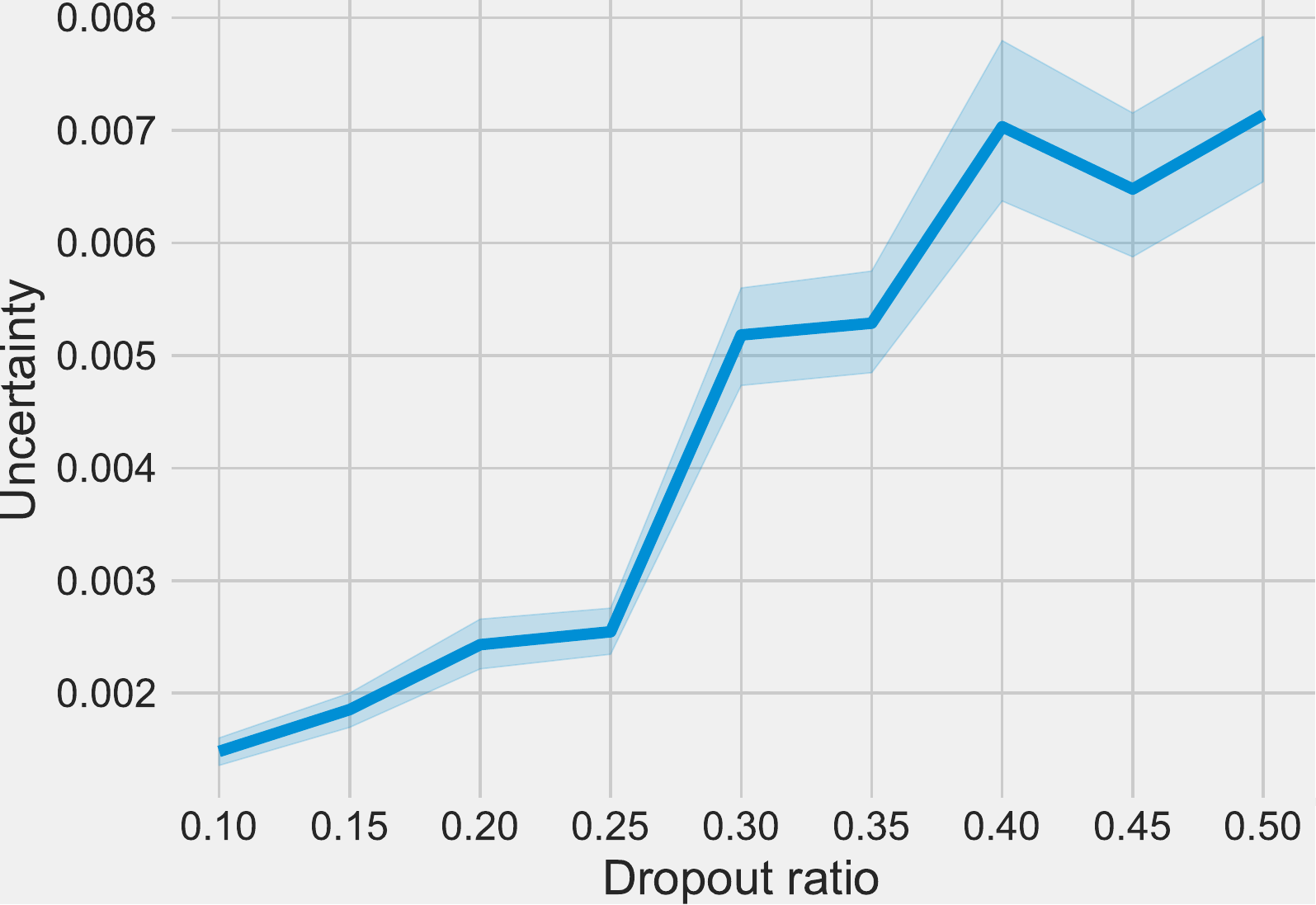}
    \caption{NEXET}
    \label{fig:ssd300-nexet-unc.pdf}
    \end{subfigure}
    \begin{subfigure}[b]{0.48\linewidth} \centering
    \includegraphics[width=0.9\linewidth]{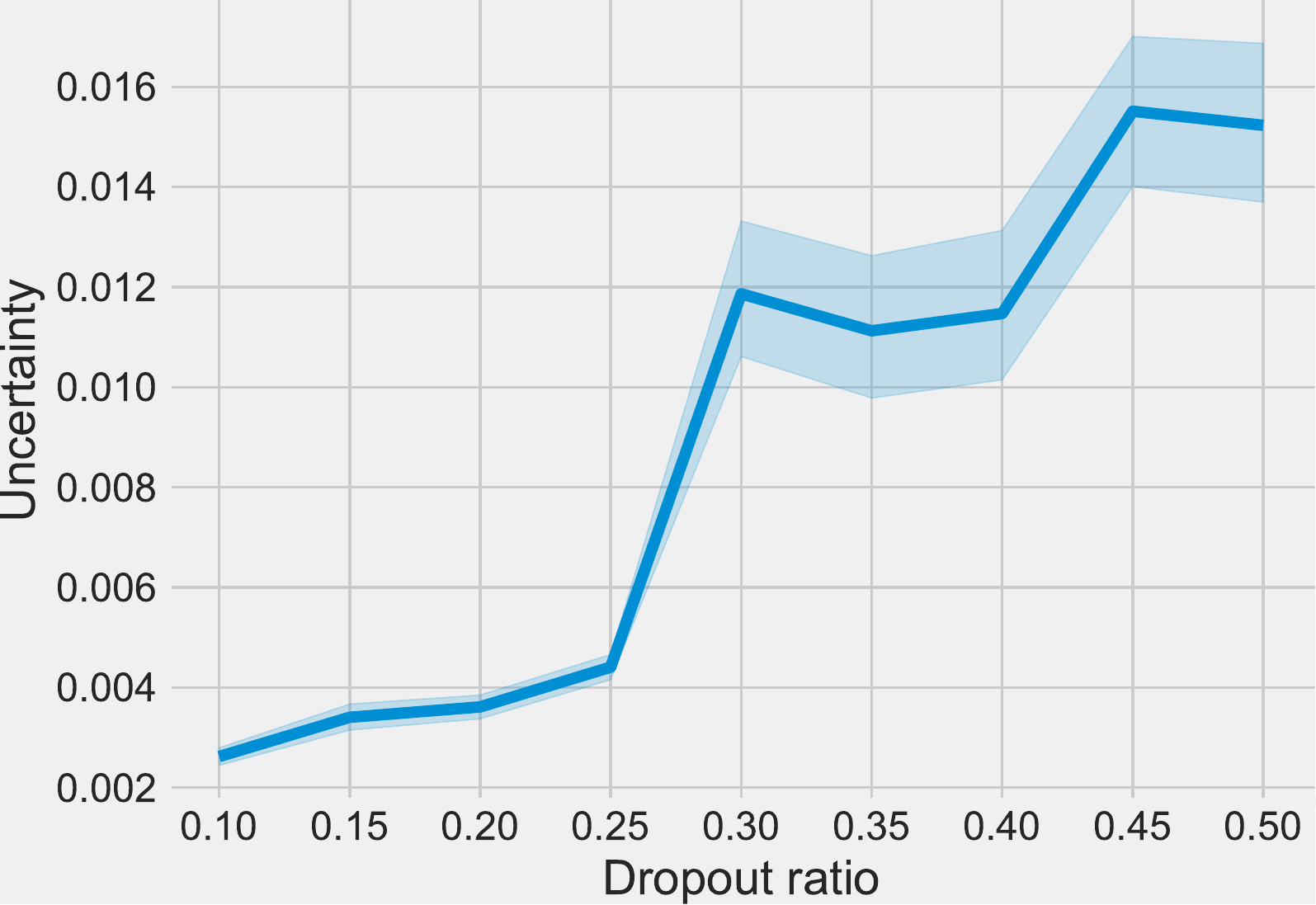}
    \caption{KITTI}
    \label{fig:ssd300-kitti-unc.pdf}
    \end{subfigure}
    \caption{SSD300 model uncertainty changes with different $p$.}
    \label{fig:ssd300-unc}
\end{figure}

\begin{figure}
    \centering
    \begin{subfigure}[b]{0.48\linewidth} \centering
    \includegraphics[width=0.9\linewidth]{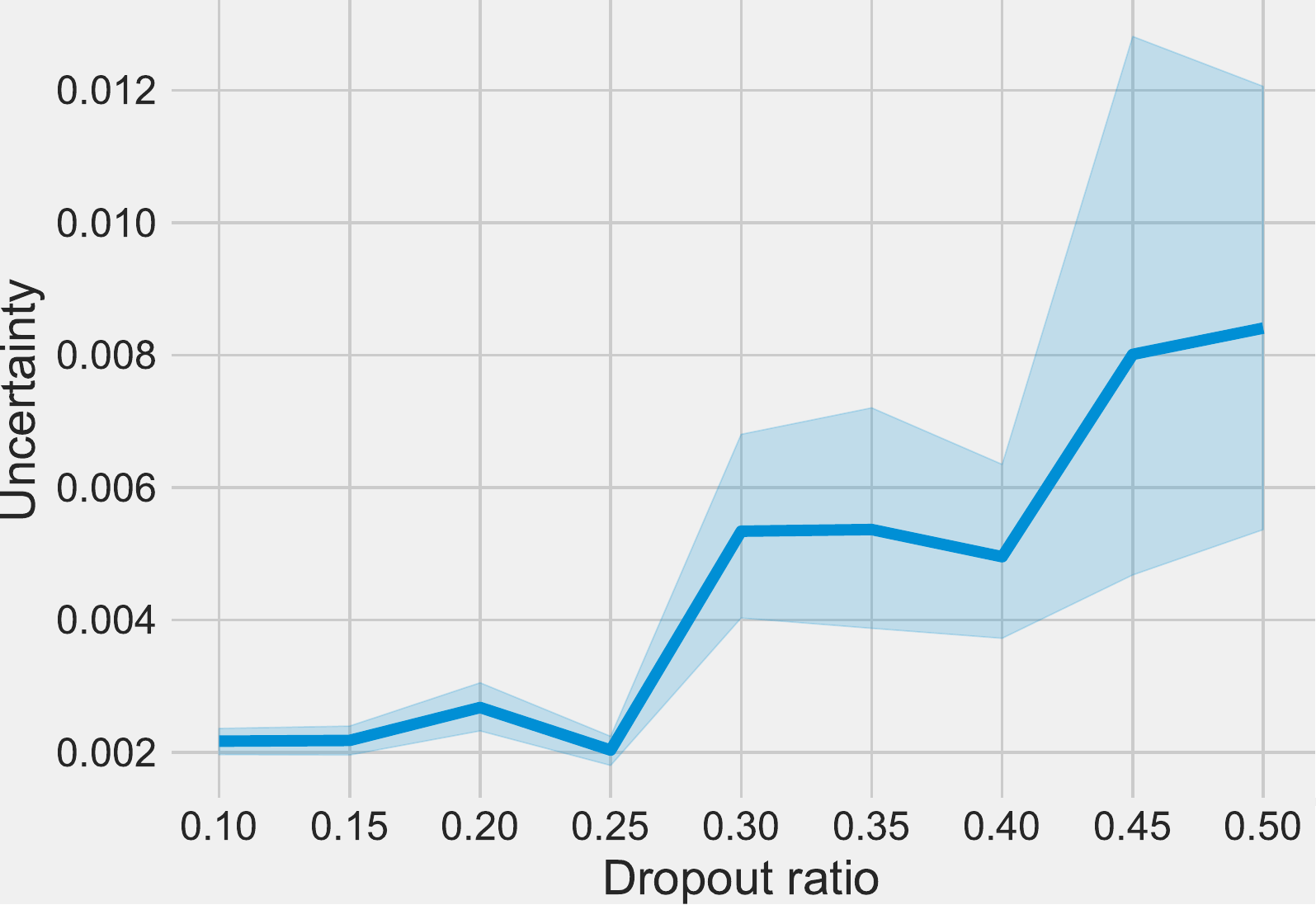}
    \caption{Stanford}
    \label{fig:yolo-stanford_cars-unc.pdf}
    \end{subfigure}
    \begin{subfigure}[b]{0.48\linewidth} \centering
    \includegraphics[width=0.9\linewidth]{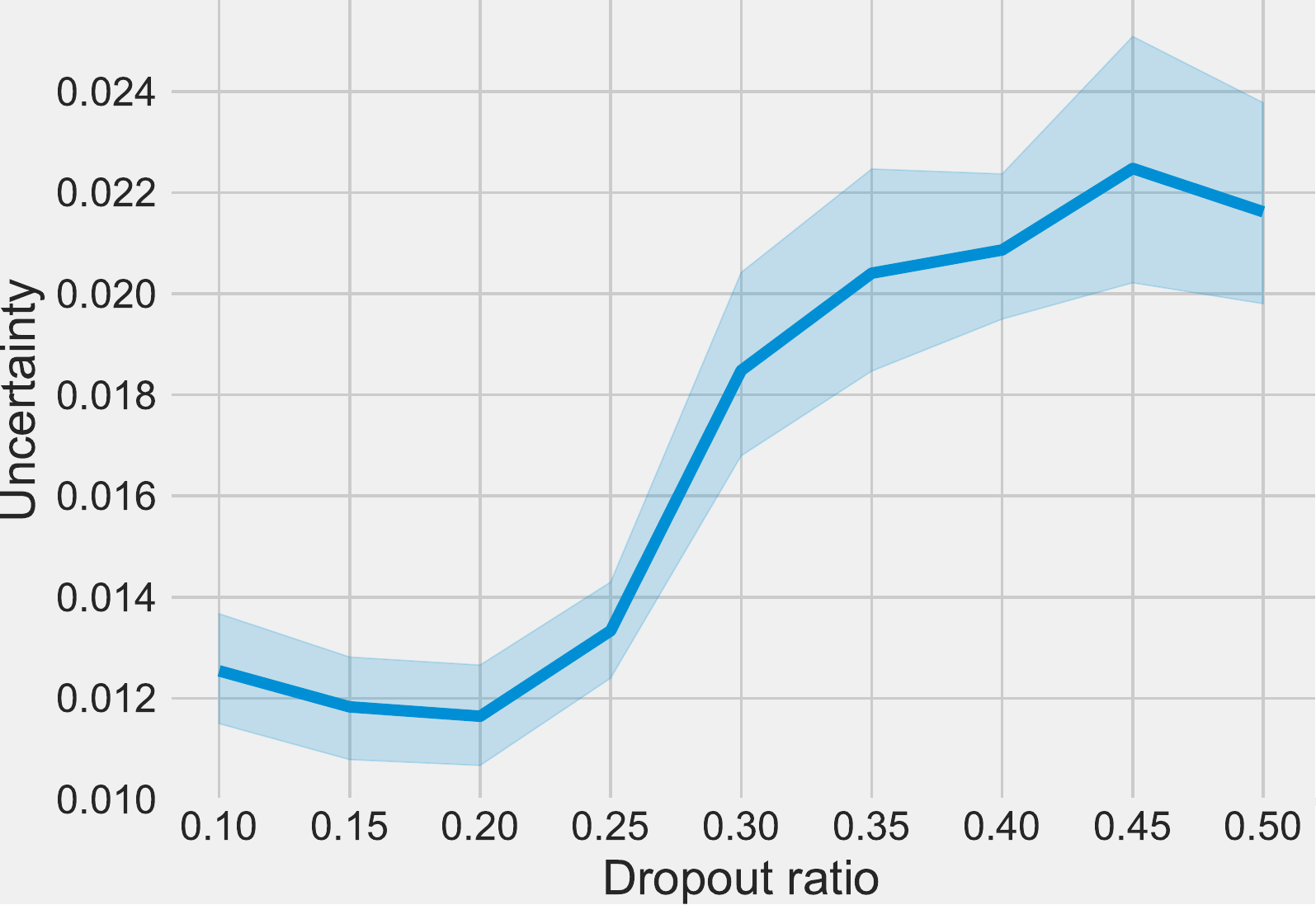}
    \caption{Berkeley}
    \label{fig:yolo-nexet-unc.pdf}
    \end{subfigure}
    \begin{subfigure}[b]{0.48\linewidth} \centering
    \includegraphics[width=0.9\linewidth]{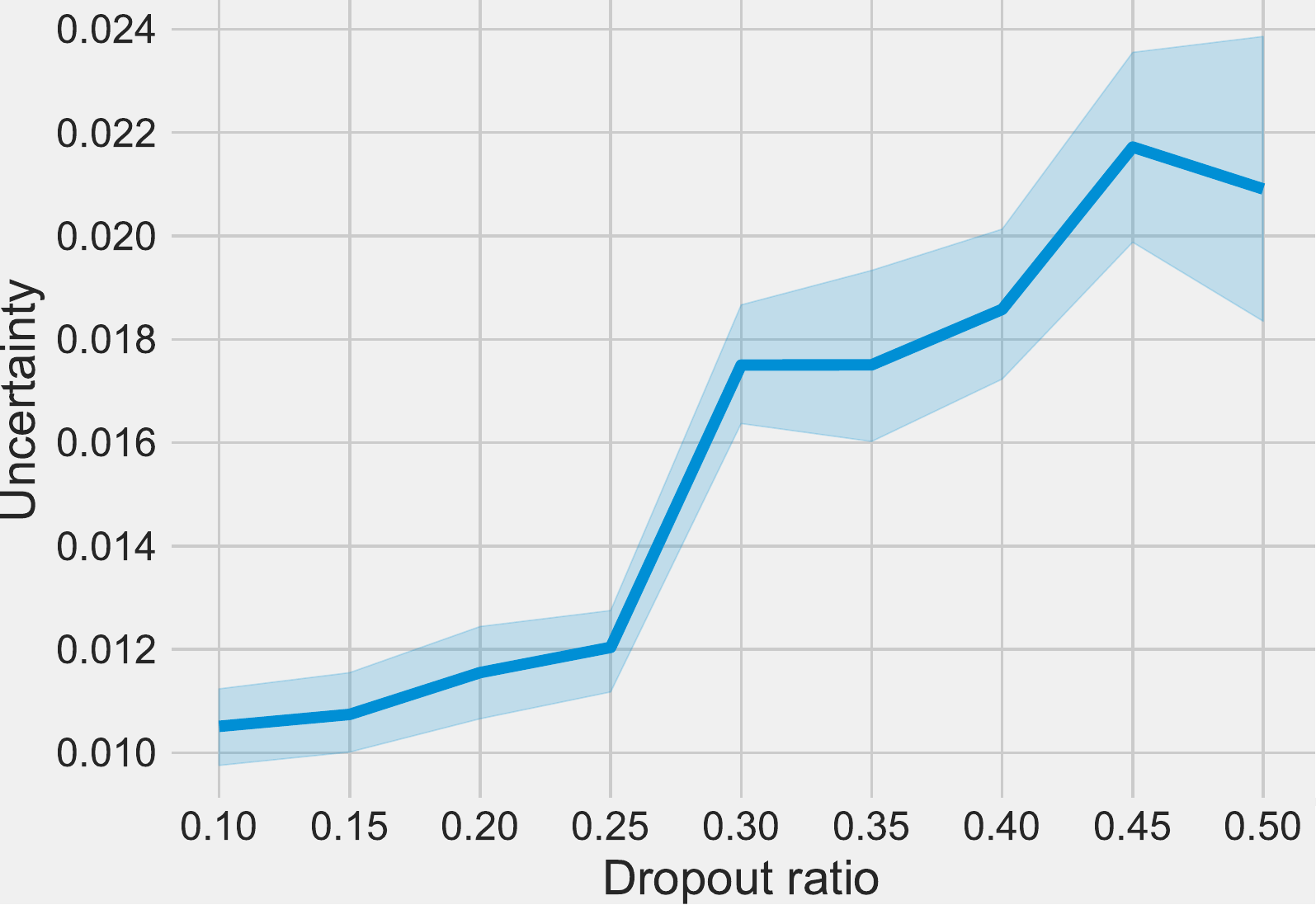}
    \caption{NEXET}
    \label{fig:yolo-nexet-unc.pdf}
    \end{subfigure}
    \begin{subfigure}[b]{0.48\linewidth} \centering
    \includegraphics[width=0.9\linewidth]{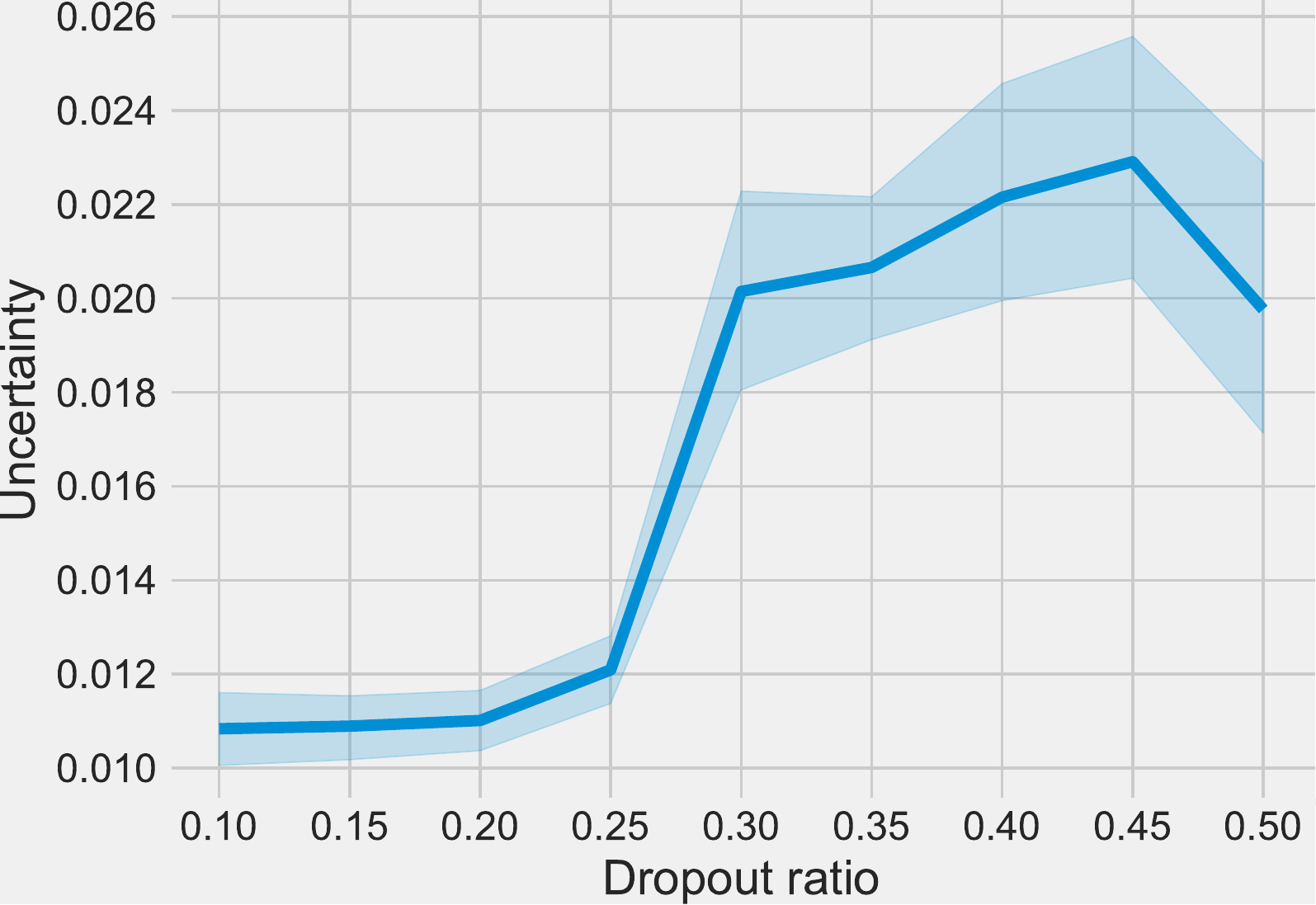}
    \caption{KITTI}
    \label{fig:yolo-kitti-unc.pdf}
    \end{subfigure}
    \caption{YoLo model uncertainty changes with different $p$.}
    \label{fig:yolo-unc}
\end{figure}

Figures \ref{fig:ssd512-unc} - \ref{fig:yolo-unc} show the $p$ and the uncertainty values. According to the figures, if an object detection model has a high dropout ratio ($p>0.3$), then the resulting uncertainty value becomes higher. The figures show that there is an increasing trend in uncertainty with an increasing $p$. 
Thus, we conclude that in order to avoid increasing the uncertainty problem caused by the MC dropout method itself, a small value $p$ should be chosen.

\subsubsection{RQ2} 


\begin{figure}[htbp!]
    \centering
    \begin{subfigure}[b]{0.48\linewidth} \centering
    \includegraphics[width=0.9\linewidth]{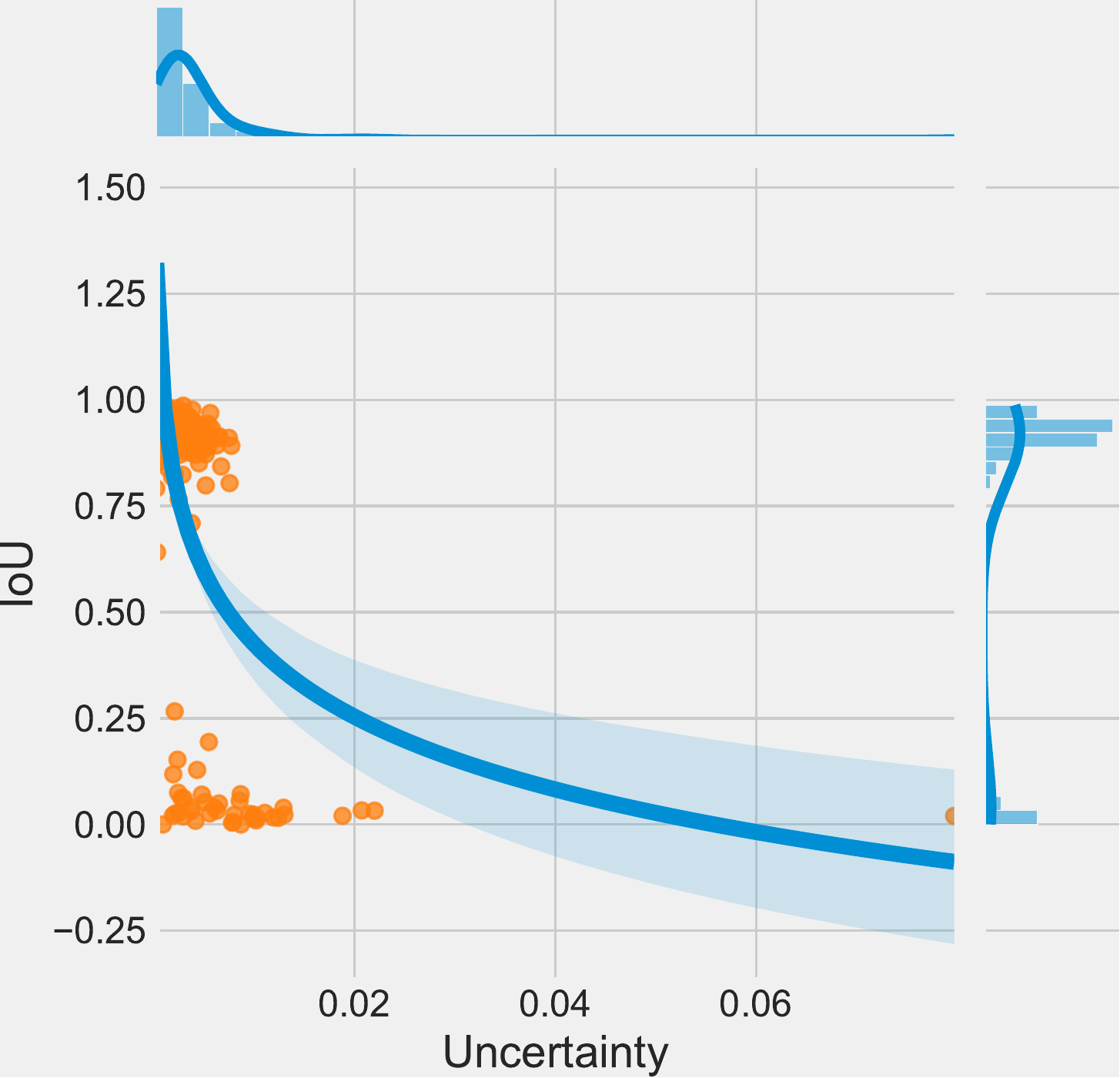}
    \caption{Stanford}
    \label{fig:ssd512-stanford_cars-perf.pdf}
    \end{subfigure}
    \begin{subfigure}[b]{0.48\linewidth} \centering
    \includegraphics[width=0.9\linewidth]{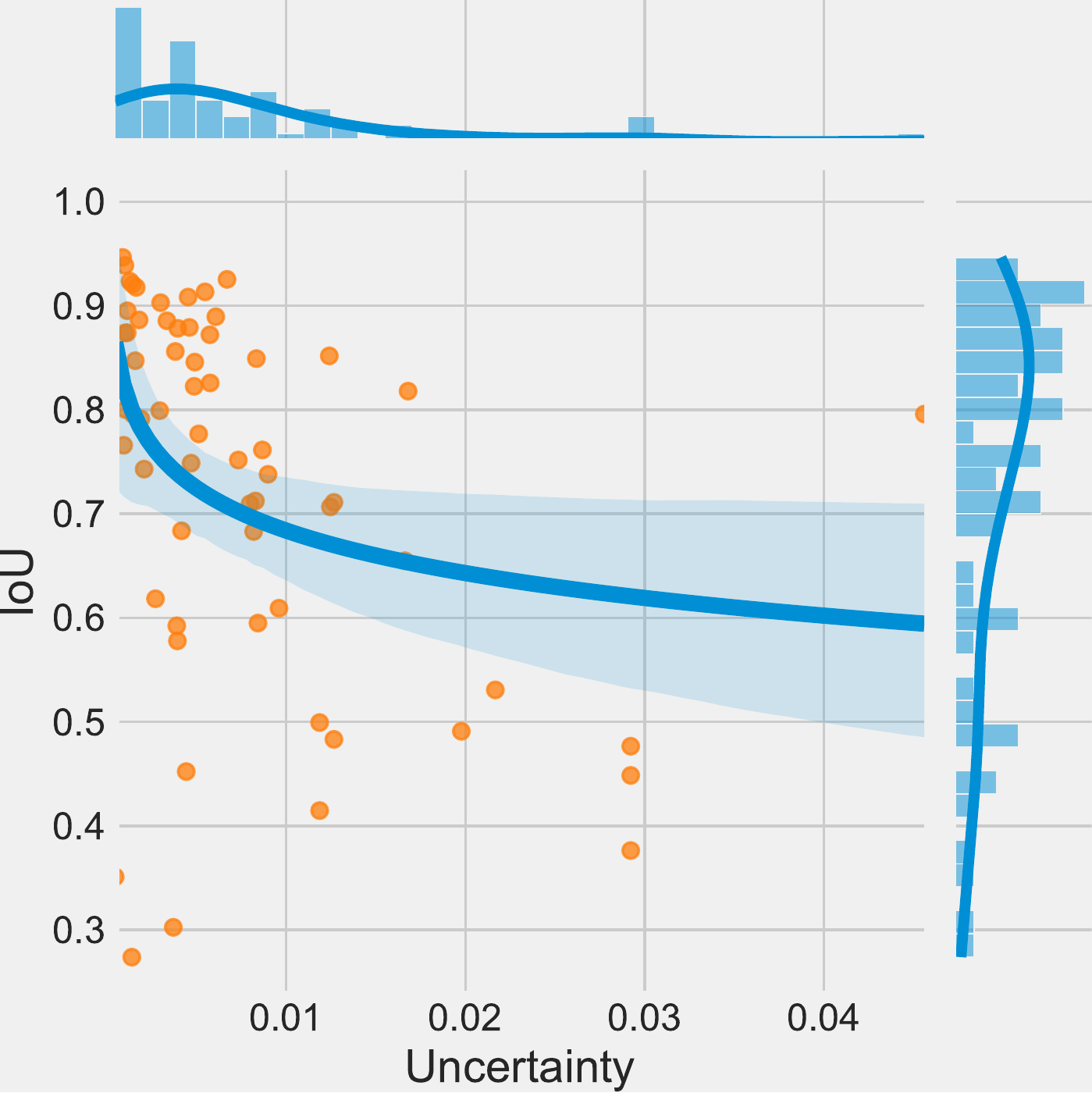}
    \caption{Berkeley}
    \label{fig:ssd512-nexet-perf.pdf}
    \end{subfigure}
    \begin{subfigure}[b]{0.48\linewidth} \centering
    \includegraphics[width=0.9\linewidth]{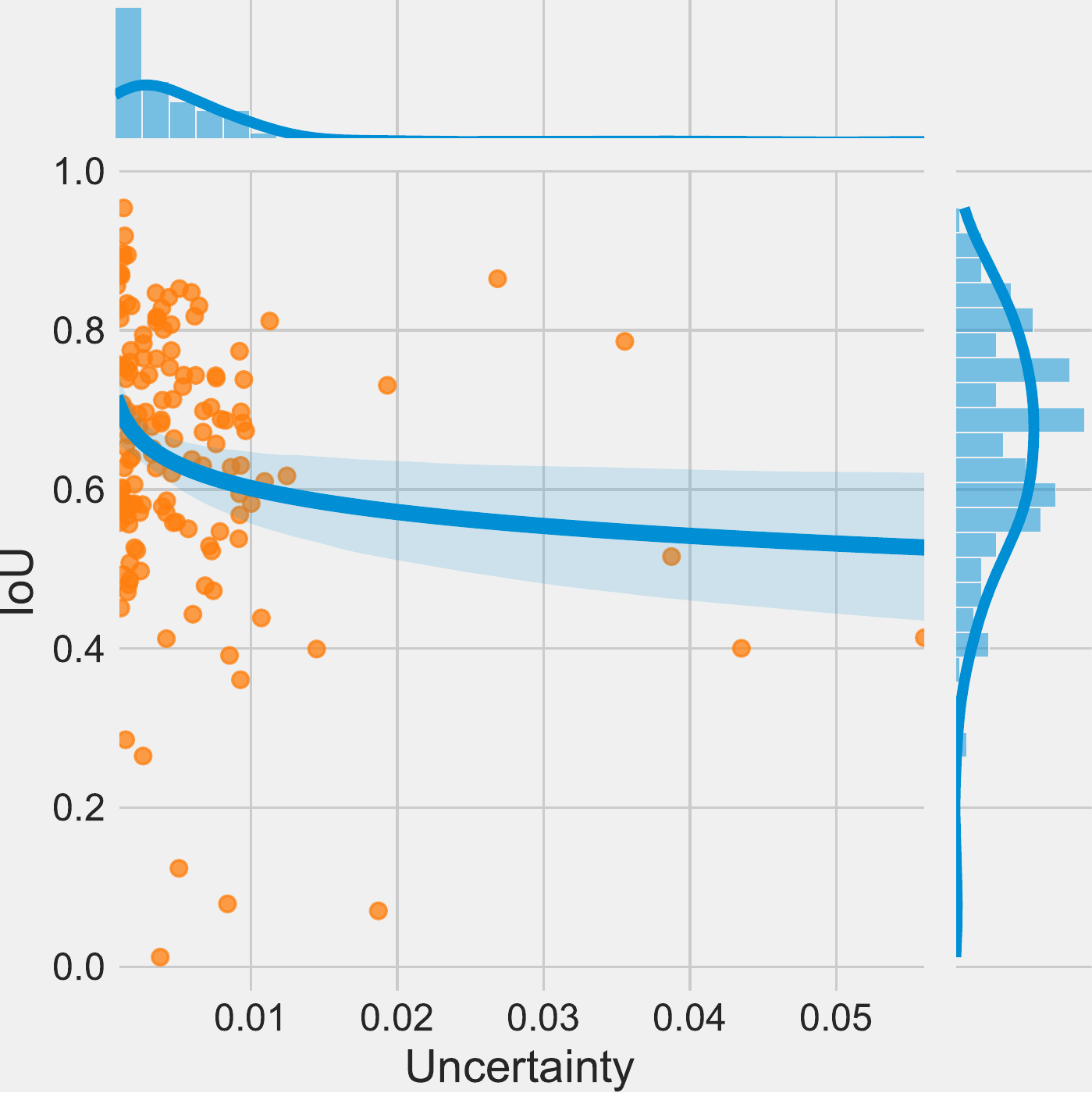}
    \caption{NEXET}
    \label{fig:ssd512-nexet-perf.pdf}
    \end{subfigure}
    \begin{subfigure}[b]{0.48\linewidth} \centering
    \includegraphics[width=0.9\linewidth]{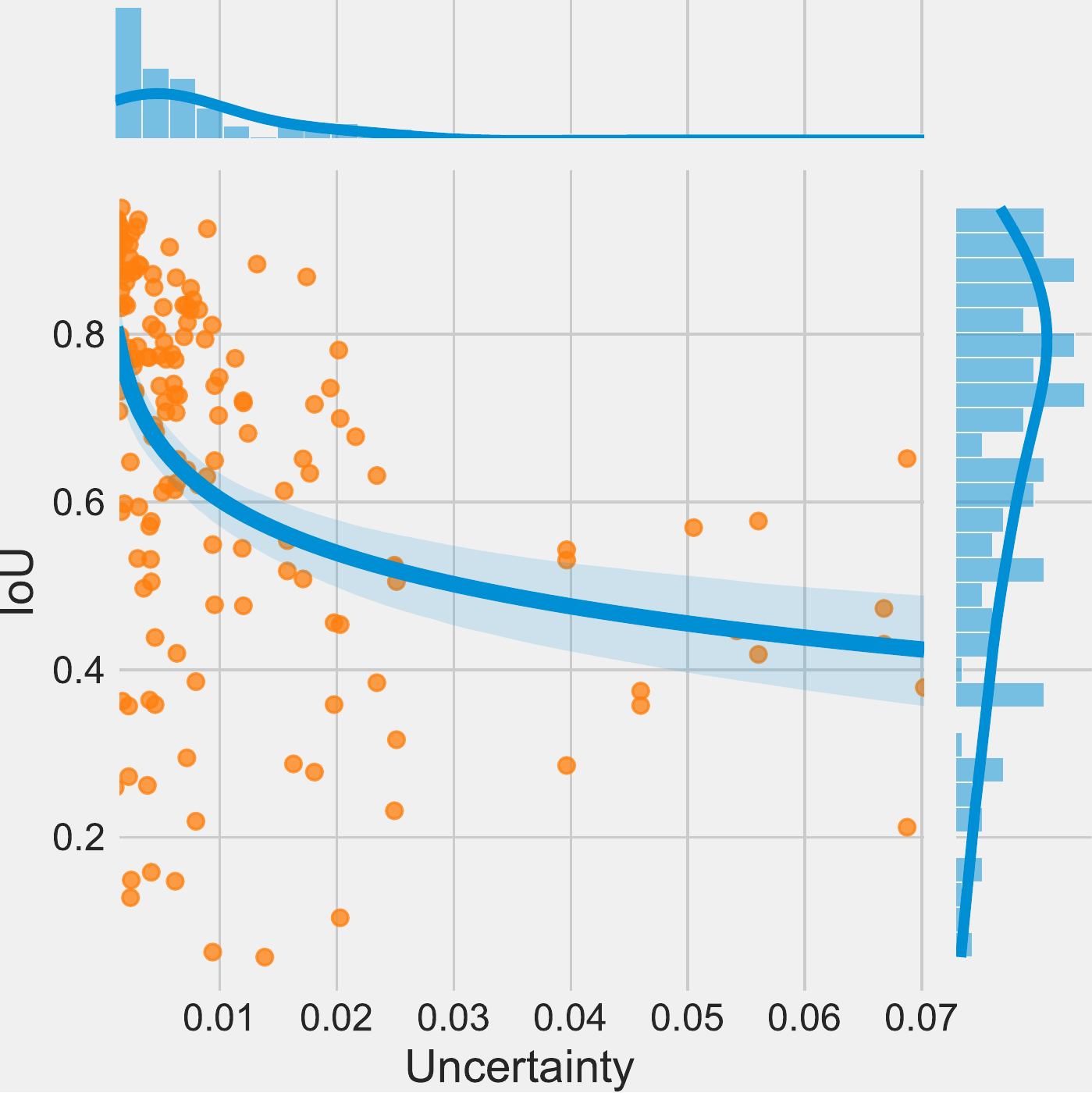}
    \caption{KITTI}
    \label{fig:ssd512-kitti-perf.pdf}
    \end{subfigure}
    \caption{SSD512 performance with different uncertainty values.}
    \label{fig:ssd512-perf}
\end{figure}

\begin{table*}

    \centering 
        \caption{Results of the Pearson correlation coefficient between prediction surface uncertainty and IoU.}
    \label{tab:results} \footnotesize
\begin{tabular}{|c|c|c|c|c|c|c|c|}
\hline
\multicolumn{2}{|c|}{} & \multicolumn{2}{|c|}{\textbf{YoLo}} & \multicolumn{2}{|c|}{\textbf{SSD300}} & \multicolumn{2}{|c|}{\textbf{SSD512}} \\ \hline
\textbf{} &  \textbf{p} & \textbf{r} & \textbf{p-val} & \textbf{r} & \textbf{p-val} & \textbf{r} & \textbf{p-val} \\ \hline
\multirow{9}{*}{\rotatebox[origin=c]{90}{KITTI}}& 0.10  & -0.2274 & 4.53e-04 & -0.2141 & 1.86e-04 & -0.3116 & 2.83e-05 \\
& 0.15 & -0.2279 & 4.79e-04 & -0.3340 & 3.34e-09 & -0.3040 & 1.63e-05 \\
& 0.20 & -0.1324 & 5.36e-02 & -0.2710 & 1.54e-06 & -0.2573 & 2.02e-04 \\
& 0.25 & -0.3241 & 5.97e-07 & -0.2726 & 1.21e-07 & -0.3848 & 3.93e-07 \\
& 0.30 & -0.2483 & 4.48e-04 & -0.2876 & 2.23e-07 & -0.3399 & 1.09e-06 \\
& 0.35 & -0.2089 & 2.01e-03 & -0.2579 & 6.21e-06 & -0.3190 & 5.21e-05 \\
& 0.40 & -0.2818 & 4.05e-05 & -0.2329 & 4.24e-05 & -0.2310 & 3.77e-04 \\
& 0.45 & -0.1650 & 2.28e-02 & -0.4131 & 1.53e-14 & -0.2127 & 4.82e-03 \\
& 0.50 & -0.2804 & 1.93e-04 & -0.2256 & 1.18e-04 & -0.4412 & 2.07e-10 \\ \hline
\multirow{9}{*}{\rotatebox[origin=c]{90}{Berkeley}} & 0.10  & -0.3743 & 2.25e-07 & -0.3778 & 2.72e-08 & -0.4940 & 6.46e-07 \\
& 0.15  & -0.3265 & 1.49e-04 & -0.4098 & 6.29e-09 & -0.5528 & 1.20e-05 \\
& 0.20  & -0.3665 & 9.57e-05 & -0.4394 & 7.52e-11 & -0.6793 & 1.90e-10 \\
& 0.25  & -0.4161 & 1.11e-05 & -0.3841 & 4.49e-07 & -0.3670 & 3.34e-03 \\
& 0.30  & -0.4247 & 4.11e-05 & -0.4356 & 1.68e-09 & -0.4724 & 4.75e-05 \\
& 0.35  & -0.4711 & 1.98e-05 & -0.3156 & 2.91e-05 & -0.3362 & 9.86e-03 \\
& 0.40  & -0.1481 & 5.99e-02 & -0.3293 & 4.70e-06 & -0.5052 & 1.10e-06 \\
& 0.45  & -0.3007 & 9.73e-03 & -0.4091 & 7.76e-08 & -0.6211 & 6.82e-06 \\
& 0.50  & -0.3564 & 6.42e-05 & -0.2122 & 1.00e-02 & -0.2589 & 1.12e-02 \\ \hline
\multirow{9}{*}{\rotatebox[origin=c]{90}{NEXET}} & 0.10  & -0.3965 & 1.92e-06 & -0.1970 & 6.42e-03 & -0.1693 & 4.62e-02 \\
& 0.15  & -0.2068 & 3.34e-02 & -0.1641 & 2.18e-02 & -0.2957 & 8.97e-04 \\
& 0.20  & -0.2943 & 1.55e-03 & -0.3278 & 1.27e-05 & -0.2876 & 5.43e-04 \\
& 0.25  & -0.2012 & 1.92e-02 & -0.2302 & 1.35e-03 & -0.2066 & 1.23e-02 \\
& 0.30  & -0.1897 & 3.32e-02 & -0.2858 & 7.00e-05 & -0.1667 & 4.58e-02 \\
& 0.35  & -0.2571 & 1.62e-02 & -0.1581 & 2.53e-02 & -0.2428 & 2.48e-03 \\
& 0.40  & -0.2989 & 2.93e-03 & -0.2438 & 9.70e-04 & -0.3189 & 1.93e-04 \\
& 0.45  & -0.3710 & 4.13e-03 & -0.2039 & 3.76e-03 & -0.3289 & 1.40e-04 \\
& 0.50  & -0.2788 & 4.11e-02 & -0.2121 & 8.47e-03 & -0.2725 & 1.03e-03 \\ \hline
\multirow{9}{*}{\rotatebox[origin=c]{90}{Stanford}}& 0.10  & -0.5002 & 1.160e-07 & -0.2807 & 6.77e-12 & -0.4197 & 1.02e-10 \\
& 0.15 & -0.6169 & 2.52e-10 & -0.2547 & 5.52e-10 & -0.5766 & 8.31e-21 \\
& 0.20 & -0.7152 & 1.02e-14 & -0.3400 & 4.40e-17 & -0.4325 & 2.11e-11 \\
& 0.25 & -0.5619 & 2.19e-08 & -0.4810 & 1.22e-34 & -0.4377 & 1.58e-11 \\
& 0.30 & -0.7161 & 5.66e-14 & -0.3713 & 3.83e-20 & -0.3506 & 9.84e-08 \\
& 0.35 & -0.7816 & 9.77e-20 & -0.3580 & 9.54e-19 & -0.3909 & 2.25e-09 \\
& 0.40 & -0.7586 & 3.04e-18 & -0.4667 & 1.89e-32 & -0.5171 & 2.62e-16 \\
& 0.45 & -0.5765 & 4.13e-09 & -0.4606 & 1.73e-31 & -0.4123 & 2.33e-10 \\
& 0.50 & -0.7769 & 1.52e-17 & -0.4475 & 1.03e-29 & -0.2846 & 1.97e-05 \\ \hline
\end{tabular}
\end{table*}

Table \ref{tab:results} shows the experimental results for all the datasets with all the object detection models. We used the Pearson correlation method to find the correlation between prediction surface uncertainty and IoU \cite{pearson1895notes}. In the table, $p$ is the dropout ratio; $r$ is the correlation coefficient; and $p$-value shows the significance of the correlation. A p-value lower than $0.05$ indicates the statistically significant correlation coefficient. We see that all the p-values are less than 0.05, and all the \textit{r} values are negative. This means that there is a statistically significant negative correlation between uncertainty and IoU. Thus, we can conclude that with the increase in uncertainty, the prediction accuracy of object detection in autonomous driving will be significantly decreased.  

For a better visualization, as an example, Figure \ref{fig:ssd512-perf} presents the scatter plots of the prediction surface uncertainty and object detection performance metric IOU, with 0.25 dropout ratio for the SSD512 model. We can see that with the increase in the prediction surface, the IoU values tend to decrease. This once again confirmed our conclusion drawn above.


\subsection{Threats to Validity}\label{sec:threats_to_val}
In our experiment, we used four datasets. More are needed to generalize the results. But, these datasets reflect different types of object detection, e.g., pedestrians, cars, and traffic signs. Another threat is the threshold value, $t$, for IoU, which we set to $0.5$. However, new experiments with varying $t$ are need to study its affect on the results. Regarding dropout ratio, $p$, to quantify the prediction uncertainty, we added the prediction-time dropouts to the last layers of the DL models. Adding dropouts to each layer might lead to unstable predictions of the models. Additional experiments are needed for further investigation. 


\section{Conclusion and Future Works}\label{sec:conclusion}
We proposed a new uncertainty quantification method called PURE (Prediction sURface uncErtainty) for object recognition described as a regression model with more than one outputs in the context of autonomous driving. PURE calculates the convex hull area of the predicted object localization generated by the MC dropout approach. Our experimental results based on three object recognition models, and four datasets suggest that PURE is an effective to calculate the object detection models' epistemic uncertainty.

In the future, we will apply the DNN testing metrics, e.g., neuron coverage, and neuron boundary coverage to assess PURE's effectiveness. These metrics test DL models to check any unnecessary layers or neurons in the model. Thus, our work will reveal the relationship between highly uncertain instances and their metrics values.


\section*{Acknowledgment}
The work is supported by Co-evolver (No. 286898/F20) funded by the Research Council of Norway under the FRIPRO program. It is also partially supported by the National Natural Science Foundation of China under Grant No. 61872182.

\bibliographystyle{IEEEtran}
\bibliography{myref}

\end{document}